%% file: main.tex
\documentclass[manuscript,arxiv]{acmart}
\usepackage[utf8]{inputenc} 
\usepackage{hyperref}       
\usepackage{url}            
\usepackage{booktabs}       
\usepackage{amsfonts}       
\usepackage{nicefrac}       
\usepackage{microtype}      
\usepackage{xcolor}         
\usepackage{amsmath}
\usepackage{tabularx}
\usepackage{multirow}
\usepackage{multicol}
\usepackage{graphicx}
\usepackage{float}
\usepackage{cleveref}
\usepackage{subcaption} 

\AtBeginDocument{%
  }

\author{Jiankai Tang}
\authornote{Equal contributions.}
\orcid{0009-0009-5388-4552}
\email{tjk24@mails.tsinghua.edu.cn}
\affiliation{%
  \institution{Key Laboratory of Pervasive Computing, Ministry of Education, Department of Computer Science and Technology, Tsinghua University; Ant Group}
  \country{China}
}

\author{Jiacheng Liu}
\authornotemark[1]
\affiliation{%
  \institution{Key Laboratory of Pervasive Computing, Ministry of Education, Department of Computer Science and Technology, Tsinghua University}
  \country{China}
}

\author{Renling Tong}

\author{Kai Zhu}

\author{Zhe Li}
\affiliation{%
  \institution{Ant Group}
  \country{China}
}

\author{Xin Yi}
\affiliation{%
  \institution{Key Laboratory of Pervasive Computing, Ministry of Education, Department of Computer Science and Technology, Tsinghua University}
  \country{China}
}
\author{Junliang Xing}
\affiliation{%
  \institution{Key Laboratory of Pervasive Computing, Ministry of Education, Department of Computer Science and Technology, Tsinghua University}
  \country{China}
}

\author{Yuanchun Shi}
\orcid{0000-0003-2273-6927}
\affiliation{%
  \institution{Key Laboratory of Pervasive Computing, Ministry of Education, Department of Computer Science and Technology, Tsinghua University}
  \country{China}
}
\affiliation{%
  \institution{Intelligent Computing and Application Laboratory of Qinghai Province, Qinghai University}
  \country{China}
}
\email{shiyc@tsinghua.edu.cn}

\author{Yuntao Wang}
\authornote{Corresponding author.}
\email{yuntaowang@tsinghua.edu.cn}
\orcid{0000-0002-4249-8893}
\affiliation{%
  \institution{Key Laboratory of Pervasive Computing, Ministry of Education, Department of Computer Science and Technology, Tsinghua University}
  \country{China}
}

\begin{document}

\title{Exploring Reliable PPG Authentication on Smartwatches in Daily Scenarios}


\begin{abstract}
Photoplethysmography (PPG) Sensors, widely deployed in smartwatches, offer a simple and non-invasive authentication approach for daily use. However, PPG authentication faces reliability issues due to motion artifacts from physical activity and physiological variability over time. To address these challenges, we propose MTL-RAPID, an efficient and reliable PPG authentication model, that employs a multitask joint training strategy, simultaneously assessing signal quality and verifying user identity. The joint optimization of these two tasks in MTL-RAPID results in a structure that outperforms models trained on individual tasks separately, achieving stronger performance with fewer parameters. In our comprehensive user studies regarding motion artifacts (N = 30), time variations (N = 32), and user preferences (N = 16), MTL-RAPID achieves a best AUC of 99.2\% and an EER of 3.5\%, outperforming existing baselines. We open-source our PPG authentication dataset along with the MTL-RAPID model to facilitate future research on \href{https://github.com/thuhci/ANT_PPG}{GitHub}. 

\end{abstract}


\begin{teaserfigure}
  \includegraphics[width=\textwidth]{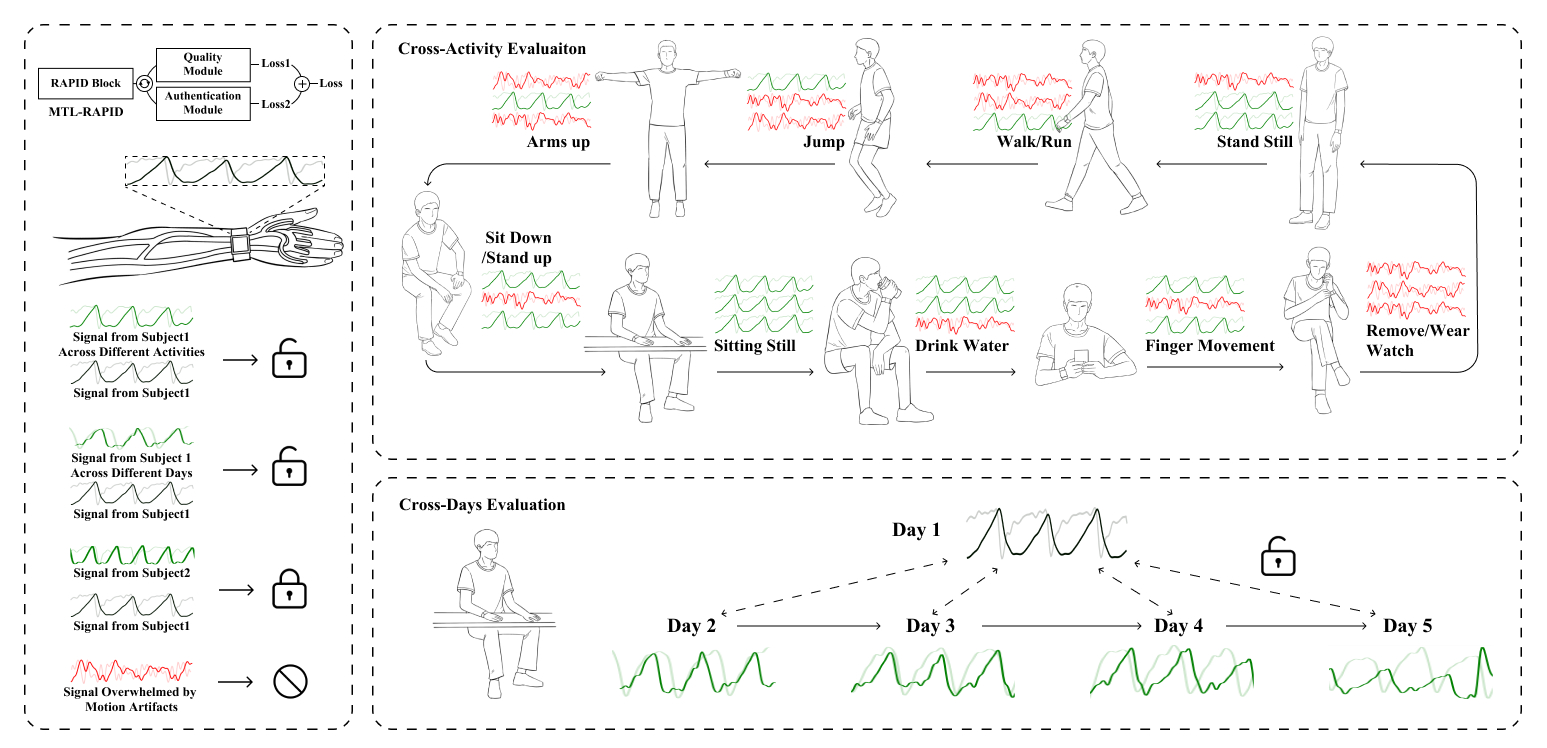}
  \caption{\textbf{MTL-RAPID makes PPG authentication on smartwatches more reliable.}  MTL-RAPID simultaneously assesses signal quality and verifies user identity with only 80k parameters. User studies were conducted to evaluate its performance under cross-activity and cross-day scenarios. }
  \label{fig:teaser}
\end{teaserfigure}


\maketitle

\input{Sections/0-intro}

\input{Sections/1-Related-Work}

\input{Sections/2-Method}
\input{Sections/3-0-Feasible-Study}

\input{Sections/3-1-Motion-Study}
\input{Sections/4-LongTerm-Study}
\input{Sections/6-Ablation-Study}
\input{Sections/5-Usable-Study}

\input{Sections/7-Discussion}

\section{Conclusion}
In this paper, we present MTL-RAPID, the first model to seamlessly integrate signal quality assessment with PPG-based identity verification. Designed as a motion-robust solution, MTL-RAPID demonstrates superior performance in handling real-world daily activities compared to existing methods. By leveraging a multi-task learning strategy, our model significantly enhances authentication accuracy, achieving remarkable results with a 99.2\% AUC and a 3.5\% EER on the challenging noisy ANT dataset, outperforming all existing baselines. Additionally, this study pioneers the evaluation of PPG authentication reliability across multiple days, effectively addressing the critical challenge of time-varying PPG signal characteristics. Our user study further validates the practicality of MTL-RAPID, revealing a statistically significant user preference (p < 0.05) for our PPG authentication method over traditional PIN-based approaches. To foster further research and innovation in PPG authentication on smartwatches, we will open-source the MTL-RAPID model along with the ANT dataset.

\bibliographystyle{ACM-Reference-Format}
\bibliography{refs}

\end{document}

%% file: Sections/0-intro.tex
\section{Introduction} \label{sec: introduction}



As smartwatches become increasingly integrated into daily life, reliable authentication is crucial for safeguarding sensitive data, such as health information, messages, and payment details. Smartwatches frequently store personal data and act as access points for connected devices or services~\cite{koushki2021smartphone, shahzad2017continuous}, making them attractive targets for unauthorized access. As their functionality expands, implementing reliable authentication methods is imperative to prevent unauthorized access and data breaches.

Traditional PIN-based authentication on smartwatches poses challenges due to small screens, slower input speeds, and usability issues during movement~\cite{oakley2018personal}. These factors also lead to security risks, as users often choose simpler PINs, making them vulnerable to shoulder surfing~\cite{Harbach:2016:ASU:2858036.2858267, vonZezschwitz:2014:HIS:2639189.2639218}. Similarly, static biometric authentication methods, such as fingerprint and facial recognition, provide user-friendly and secure alternatives but require additional sensors, which can increase the size and weight of smartwatches, reducing their convenience. 
Bluetooth-based authentication, though convenient, introduces proximity-based risks such as interception or spoofing~\cite{Seneviratne2017Challenge}. These challenges highlight the need for adaptive, secure, and user-friendly authentication solutions specifically designed for wearable devices.

Photoplethysmography (PPG), a commonly used technique in smartwatches for health monitoring, offers a promising alternative for authentication. PPG measures blood volume changes in the microvascular bed of tissues and can extract physiological signals such as heart rate and blood oxygen levels. More importantly, PPG data reveals distinctive physiological patterns unique to each individual, enabling its use for biometric authentication, commonly referred to as PPG authentication ~\cite{choudhary2016robust}. This method does not require additional hardware, making it highly practical and cost-effective for broad adoption. By analyzing PPG data, a smartwatch can provide non-invasive, efficient authentication, enhancing both security and user convenience~\cite{sarkar2016biometric}.

However, despite its potential, the reliability of PPG authentication in real-world scenarios remains a challenge. Daily activities involving significant hand movements or physical exercise can introduce motion artifacts that can affect the accuracy and reliability of the authentication process. Moreover, managing power consumption is critical for smartwatch authentication design. Therefore, efficient PPG authentication algorithms are necessary to allow frequent authentication without frequent recharging~\cite{Das2018Efficient}, ensuring security while maintaining usability and appeal for everyday wear. These issues underscore the need for efficient, reliable authentication methods capable of managing variability and ensuring consistent performance even in dynamic conditions~\cite{sarkar2016biometric, shang2019usable}. 

In this paper, we explore efficient and reliable PPG on smartwatches for daily scenarios. Our approach proactively detects suitable conditions for authentication and performs the authentication task without requiring active unlocking at the moment of use, allowing the user seamless and uninterrupted access. To achieve this, we proposed the \textbf{M}ulti-\textbf{T}ask \textbf{L}earning \textbf{R}eliable and \textbf{A}ccurate \textbf{P}PG-based model for efficient \textbf{ID} authentication (MTL-RAPID), specifically designed to address the challenges of PPG-based authentication for daily scenarios. Our MTL-RAPID model utilizes a multi-task joint architecture to extract and analyze global and local vascular features, heartbeat patterns, and human motion forms from PPG signals, along with their interrelationships. By adjusting the passing rate of the quality assessment task, we can adapt authentication strength to various scenarios. With only 80k parameters, our method offers efficient and accurate quality assessment and identity verification, making it ideal for deployment on smartwatches. Our main contributions are as follows:

\begin{enumerate}
    \item We introduced MTL-RAPID, a lightweight model designed to assess PPG signal quality and perform authentication simultaneously, enabling reliable smartwatch authentication in daily scenarios.
    \item We conducted three user studies to evaluate the reliability of the MTL-RAPID model on cross-activity (N = 30), cross-day (N = 32) and user preference (N = 16), achieving 99.2\% AUC and 3.5\% EER at best, with significant user preference over PIN methods. 
    \item We open-sourced a PPG authentication dataset spanning multiple daily activities, along with the MTL-RAPID model, to enhance reproducibility and encourage the widespread adoption of PPG-based authentication (see supplementary materials).
\end{enumerate}

%% file: Sections/1-Related-Work.tex
\section{Related Work} \label{sec: Related Work}
We first reviewed existing authentication methods on smartwatches from traditional unlock patterns to new biometric approaches. We then discuss the reliability of those authentication methods in real-life mobile scenarios. At last, we investigated unsupervised and supervised PPG authentication methods.
\subsection{Authentication Methods on Smartwatches}
Explicit authentication methods, such as PINs and graphical patterns, though widely used, are plagued by significant usability issues due to the small size of smartwatch screens, necessitating precise inputs \cite{oakley2018personal}. These methods also suffer from security vulnerabilities; simplistic and memorable choices by users make these methods prone to attacks, with an attacker potentially cracking 4.6\% of 4-digit PINs within 10 online guesses \cite{markert2020pin}. Pattern authentication is similarly risky, with crack rates varying from 13.33\% to 32.55\% in different studies \cite{cha2017boosting,cho2017syspal}, influenced by well-documented biases \cite{Munyendo2021blocklist}. ~\citet{10.1145/3264929} proposed an authentication method when users write signatures, achieving an EER of 2.36\% and an AUC of 98.52\% (N = 66). These explicit methods have shown promising performance but require additional effort from users.

Implicit biometric authentication presents an alternative by using physiological or behavioral traits, which show promise in improving security without the cumbersome input methods. ~\citet{cornelius2014wearable} achieved a 13.1\% Equal Error Rate (EER) using on-wrist bioimpedance (N = 8), highlighting its potential for secure authentication.  Low-frequency vibration responses measured by ~\citet{lee2021usable} initially showed a promising 1.37\% EER (N = 19), but a rise to 4.99\% FRR over a week suggests issues with long-term stability. Acoustic response with retraining classifiers utilized by ~\citet{huh2023wristacoustic} achieves 0.79\% EER on recall-session study (N = 20). While these methods demonstrate effectiveness, they either require additional sensors or have been validated only on small sample sizes.

\label{related: PPG Watch}
PPG authentication on smartwatches presents a viable alternative due to its widespread use and low power consumption compared to other biometric methods. ~\citet{shang2019usable} were the first to explore the feasibility of combining PPG with gesture-based authentication, achieving a 91.6\% true rejection rate using wrist-worn sensors across 12 subjects. In a separate study, ~\citet{zhao2020trueheart} reported a 90.7\% accuracy using a gradient boosting tree method for PPG-based authentication on smartwatches with 20 participants across hours in a single day. Differing from these initial approaches, our research expands the field by conducting user studies with more participants with longer time intervals between sessions to address real-world challenges. 

\subsection{Reliable Authentication in Mobile Scenarios}
Reliable authentication, particularly in mobile environments, has become a critical focus in the security domain. Traditional methods such as passwords and explicit biometric verification(e.g., face~\cite{papavasileiou:2017:GCA:3204094.3204151, 7791155}, fingerprint) are known for their high reliability but often interrupt the user's workflow. In contrast, implicit authentication methods aim to enhance user experience by seamlessly integrating into daily activities(e.g., pick-up~\cite{Lee_2017}). These methods analyze subtle behavioral features such as gait patterns~\cite{5570993}, touch posture~\cite{zhang2015touch, 10.1007/978-3-319-20678-3_32}, and voice identification~\cite{7358748} to verify a user's identity without their active participation. 

Despite the appeal of implicit methods for their unobtrusive nature, they often fall short in reliability when compared to traditional authentication methods. The primary challenge is their unstable performance in uncontrolled environments, leading to redundant and potentially frustrating authentication checks~\cite{7503170, patel2016continuous, mahfouz2017survey, gonzalez2019leveraging}. Addressing these reliability issues is essential for their potential widespread commercial adoption and the overall enhancement of security in mobile scenarios.

\subsection{Photoplethysmography (PPG) Authentication Methods}

Photoplethysmography (PPG) authentication has been extensively explored, though most studies have focused on feasibility studies using fingertip PPG datasets~\cite{MIT-BIH, karlen2010capnobase, koelstra2011deap, luque2018end, sancho2018biometric,pimentel2016bidmc,lee2011MIMIC-II}. Only a few have tested methodologies on wrist-worn devices~\cite{shang2019usable,zhao2020trueheart} as detailed in section \ref{related: PPG Watch}.

Unsupervised algorithms in PPG authentication typically involve extracting features from both the time domain (e.g., peaks, intervals, slopes) and the frequency domain of the PPG signal. Methods such as peak detection, cross-correlation, Continuous Wavelet Transform (CWT), Principal Component Analysis (PCA), K-Nearest Neighbors (KNN), Linear Discriminant Analysis (LDA), and Naive Bayes classifier (NBC) are employed to cluster and identify patterns without relying on previously labeled data~\cite{choudhary2016robust, karimian2017non, sancho2018biometric, yang2021study}.

On the other hand, supervised algorithms use labeled classes to learn discriminative features crucial for authentication. Techniques include convolutional neural networks (CNNs), long short-term memory (LSTM), auto encoder (AE) and others. ~\citet{luque2018end} introduced an end-to-end network, ~\citet{pu2022novel} utilized an auto encoder to transform PPG into latent space, and ~\citet{wan2024deep} developed a deep CNN model that extracts features from both time-domain and frequency-domain signals. Despite their promising results, these methods exhibit limitations, such as the use of excessively long segments (45-60 seconds) for authentication, which is impractical for everyday use~\cite{yadav2018evaluation}. Unfortunately, some studies lack clarity regarding dataset selection, testing procedures, segment length, the number of segments evaluated, and whether segments span different subjects.

To overcome these limitations, we have organized and open-sourced our code and datasets (see supplementary materials), enhancing transparency and enabling fair comparative analysis with the state-of-the-art algorithms~\cite{ismail2020inceptiontime, cornet, wan2024deep}. To the best of our knowledge, we are the first to introduce the multi-task architecture based on the PPG optical principle for authentication, significantly enhancing efficiency and reliability under variable conditions.

%% file: Sections/2-Method.tex
\section{\textbf{R}eliable and \textbf{A}ccurate \textbf{P}PG-based model for efficient \textbf{ID} authentication (\textbf{MTL-RAPID)}}

In the methodological section of this paper, we introduce Multi-Task Learning \textbf{R}eliable and \textbf{A}ccurate \textbf{P}PG-based model for efficient \textbf{ID} authentication (\textbf{MTL-RAPID)}. We first present our optical basis in Section \ref{sec: optical} and describe the architecture of the basic RAPID block in Section \ref{sec: basemodel} and MTL-RAPID model in Section \ref{sec: MTL model}. We justify the algorithmic principles of PPG that guide our model design in Section \ref{sec: justify}. Based on the MTL-RAPID, we introduce the authentication procedure (i.e., registration and authentication) in Section \ref{sec: auth procedure}. We also included the preprocess and evaluation setup in Section \ref{sec: preprocess} and Section \ref{sec: study1 met}.

\subsection{Optical Principle}\label{sec: optical}

\begin{figure*}[htp]
  \includegraphics[width=250pt]{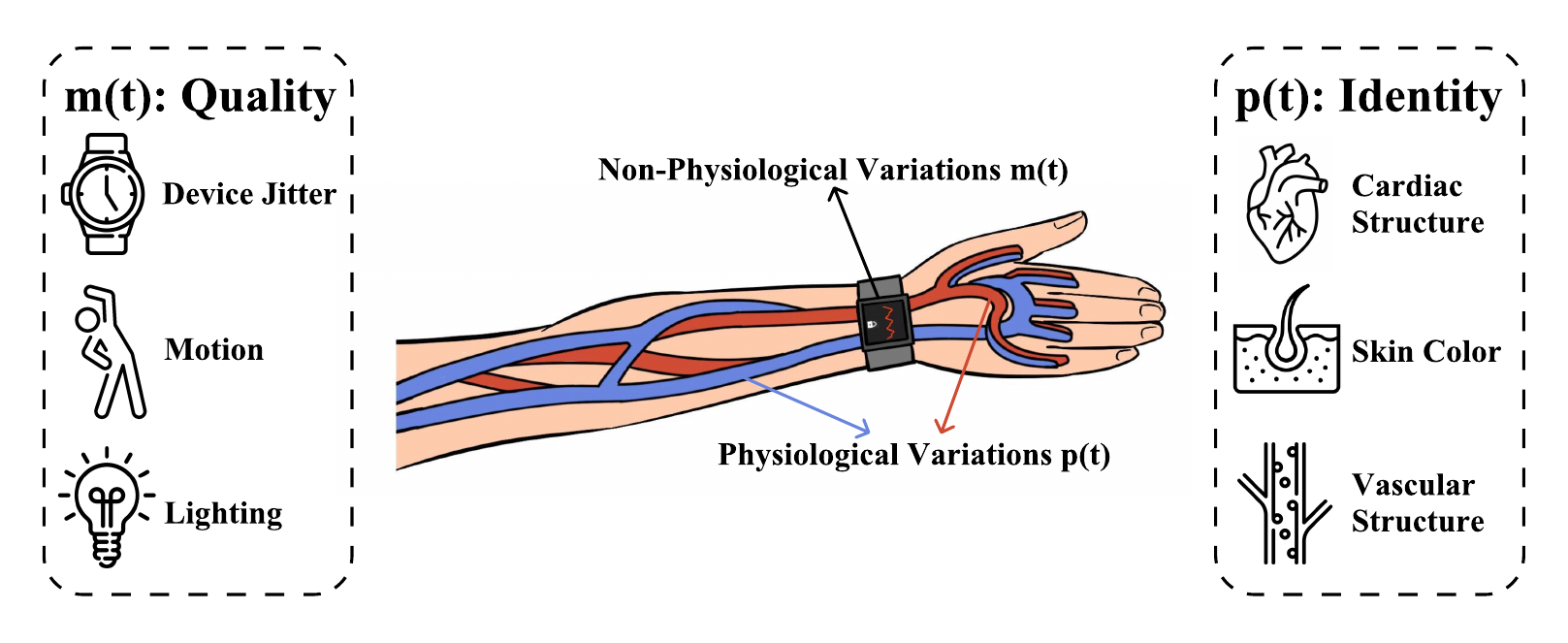}
  \caption{\textbf{Optical Principle.} PPG signals contain two key components: non-physiological variations \(m(t)\), influenced by external factors like lighting, body movements, and jitter, and physiological variations \(p(t)\), driven by heart activity, vascular structure, and skin anatomy.}
  \label{fig: optical}
\end{figure*}

Our foundational optical model uses Shafer's Dichromatic Reflection Model~\cite{shafer1985using} to analyze PPG signals. Our goal is to accurately extract vascular features, heartbeat patterns, and human motion forms along with their interrelationships from PPG, thereby enabling identity verification and quality assessment. We consider the PPG values captured by the photodetector sensor represented as follows:


\begin{equation} \label{eq:4}
	\pmb{C}_k(t)\approx 
	\pmb{u}_s \cdot I_0 \cdot \Phi(m(t),p(t))+\pmb{u}_p \cdot I_0 \cdot p(t)+\pmb{v}_n(t) + I_{ds} 
\end{equation}
%

In Eqn.~(\ref{eq:4}), $I_0$ represents the luminance intensity level,  relatively constant in wearable devices. $\pmb{u}_d$ denotes the unit color vector of skin tissue, while $\pmb{u}_p$ indicates the relative pulsatile strengths influenced by the absorption properties of hemoglobin and melanin. $p(t)$ captures the physiological changes, including individual-specific arterial characteristics and heartbeat patterns, which are fundamental to identity verification tasks based on PPG signals. The variable $m(t)$ encompasses all non-physiological variations, such as fluctuations in lighting, body movements, and the jitter of wearable devices. $\pmb{v}_n(t)$ accounts for the quantization noise inherent in the sensor, and finally $I_{ds}$ denotes other components that can be seen as constant.

Due to the entanglement between $m(t)$ and $p(t)$, $m(t)$ can interfere with the extraction of $p(t)$ features. Therefore, $m(t)$ contains quality information of the PPG signal and can be used to assess whether a specific PPG signal is suitable for identity verification tasks. Thus, it is crucial to use a method that can simultaneously extract features of $m(t)$ and $p(t)$, as well as their interactions, for reliable PPG-based identity verification.

\subsection{RAPID Block}\label{sec: basemodel}


The RAPID block is a module we designed for PPG identity verification and waveform quality assessment tasks. Inspired by the Gao et al. MTL network NDDR-CNN~\cite{gao2019nddr}, we designed separate paths within the block to extract features for each task. We then merge the features from both tasks through concatenation, enabling multi-task training. 

\begin{figure*}[htp]
  \includegraphics[width=\textwidth]{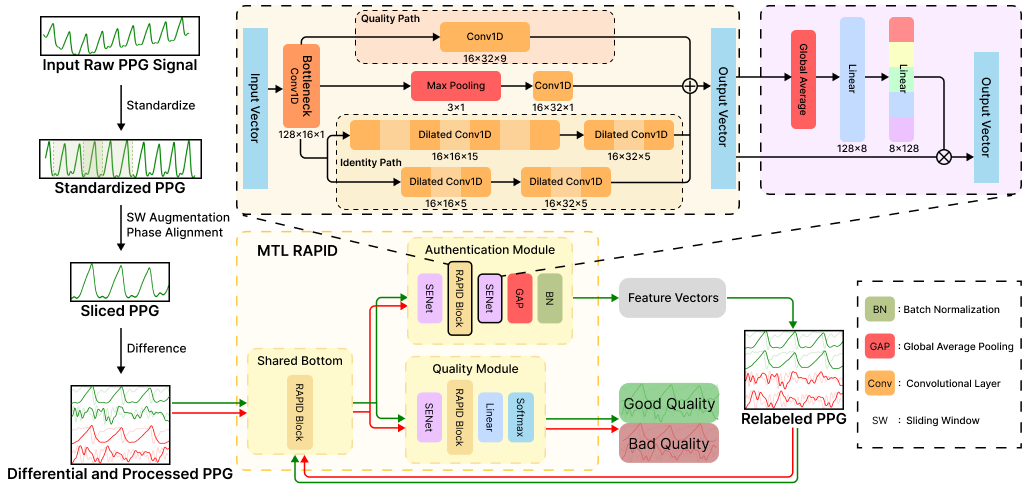}
  \vspace{-0.5cm}
  \caption{\textbf{MTL-RAPID architecture.} We propose a multi-task RAPID model and switch training procedure to accomplish waveform quality assessment and identity verification tasks simultaneously.}
  \label{fig:MTL pipeline}
\end{figure*}

As illustrated in the RAPID block shown in Figure \ref{fig:MTL pipeline}, it consists of two main components: the Quality Path and the Identity Path. The Quality Path is better suited for extracting local quality information from PPG signals. These quality features manifest as signal distortion and waveform chaos, typically due to motion and device jitter, and correspond to the \( m(t) \) component in Eqn.~(\ref{eq:4}). The Identity Path, due to its larger receptive field, is more suitable for extracting temporal features rich in user identity information, reliably learning the \( p(t) \) component in Eqn.~(\ref{eq:4}).

To streamline our model while retaining its architecture, we incorporated a bottleneck layer inspired by the InceptionTime model~\cite{ismail2020inceptiontime}, reducing the input vector's dimension and model complexity. Finally, we integrate SENet~\cite{hu2018squeeze} to introduce an attention mechanism to help the model choose important features automatically.

\subsection{MTL-RAPID}\label{sec: MTL model}

In practical applications, noise in the data can significantly impact the accuracy of authentication models. Therefore, it is crucial to first filter the signals using a waveform quality classifier.

Our MTL model comprises a shared-bottom model consisting solely of three identical RAPID blocks. Inspired by the HyperFace architecture~\cite{ranjan2017hyperface}, we utilize one shared RAPID block to extract various shallow features from PPG waveforms, which apply to both tasks. Additionally, the shared module is responsible for learning the complex interrelationships between waveform quality and physiological features, represented by the function $\Phi$ in Eqn.~(\ref{eq:4}). However, due to the significant differences in the features required for waveform quality assessment and identity verification tasks, our MTL model does not have a large shared module, and simple task-specific sub-networks consist only of a few simple linear layers, like most image recognition MTL models~\cite{ranjan2017hyperface, ma2018modeling}. Consequently, we have configured both task-specific sub-networks to be identical RAPID blocks.

\subsubsection{\textbf{Mode switch}}
The task of PPG waveform quality assessment and identity verification in our system follows a sequential relationship. Specifically, the PPG signal first undergoes waveform inspection to filter out motion samples, after which the remaining samples are subjected to identity verification. This sequential relationship distinguishes itself from the parallel relationship commonly seen in image recognition tasks, where multiple advanced features are extracted simultaneously from the same image~\cite{ranjan2017hyperface, ma2018modeling}. Due to the unidirectional flow of information in serial tasks, which requires altering the flow of information depending on the task being performed, we have implemented a mode switch following the shared module. This switch controls whether the information enters the quality assessment or the identity verification module.

\subsubsection{\textbf{MTL training approach}}
\label{sec: switch training}
Noise is a common issue in wrist PPG datasets, which can negatively impact the training of identity verification models. Although the participants' movement states are labeled during data collection, different movement states may not necessarily indicate the suitability of the signals for identity verification tasks due to possible physiological changes. Therefore, to address the absence of suitable labels for training quality assessment modules within the dataset, we employ an MTL training approach with a label correction mechanism. This method simultaneously trains both the quality assessment and identity verification modules. We use the identity verification module's ability to authenticate correctly as a standard; during training, we reassign quality labels to signal segments based on this criterion.

\subsection{Justify MTL RAPID Design for PPG Authentication}
\label{sec: justify}
In designing of our multi-task learning (MTL) model, we prioritize two crucial factors: reliability and efficiency. Authentication scenarios can range from typical daily interactions to situations demanding high security. To accommodate this variability, our model incorporates an adjustable threshold within the quality assessment module during the authentication process. This adaptability allows us to finely balance authentication strength with usability, tailoring the security measures to fit the context of use.

Efficiency is achieved by integrating multiple tasks within a single model framework, rather than maintaining separate models for each task. This unified approach not only simplifies the system architecture but also reduces the computational overhead involved in running parallel models. By consolidating tasks, our MTL model enhances processing speed and resource utilization, making it significantly more efficient for future deployment in practical applications where both performance and power consumption are critical considerations.

\subsection{Authentication Procedure}
\label{sec: auth procedure}
The PPG authentication system facilitates natural user authentication through passive interaction, removing the need for active user operations such as input or touch. The authentication process consists of two essential steps: registration and authentication.

\begin{enumerate}
  \item \textbf{Registration:} Users record at least one 6-second PPG segment as their authentication template. 
  \begin{itemize}
    \item \textit{Motion Study (Section \ref{sec: study 1}) and Cross-day Study (Section \ref{sec: study 2}):} A random 6-second PPG segment is selected from the database as the registered template to evaluate the authentication methods' effectiveness. 
    \item \textit{Usability Study (Section \ref{sec: study 3}):}  Users manually record PPG data, starting with 10 seconds. If no 6-second segment passes the quality check, recording continues until a valid segment is extracted.
  \end{itemize}

  \item \textbf{Authentication:} The system compares the recorded PPG segments with existing templates to determine user identity based on segment similarity.
  \begin{itemize}
    \item \textit{Motion Study (Section \ref{sec: study 1}) and Cross-day Study (Section \ref{sec: study 2}):} PPG segments (excluding registered templates) are randomly selected as positive pairs, while segments from other users are randomly selected as negative pairs.
    \item \textit{Usability Study (Section \ref{sec: study 3}):} Once worn, the smartwatch automatically collects PPG signals and unlocks if a 6-second segment passes the quality filter and identification process. Segments that fail the filter do not trigger any state change until a static PPG segment is authenticated.
  \end{itemize}
  \end{enumerate}



\subsection{Preprocess}
\label{sec: preprocess}
The workflow for data preprocessing and signal analysis is illustrated in Figure \ref{fig:MTL pipeline}. We first standardized the signals by resampling them to 60Hz. A 4th-order Butterworth filter with a frequency range of 0.5 Hz to 10 Hz was applied to ensure optimal signal clarity. To improve robustness, we incorporated a sliding window technique for data augmentation. A 6-second window with a 2-second overlap is used, starting at the first trough detected in the PPG signal to ensure phase alignment. Each signal is processed in both differentiated and non-differentiated forms before being input into the system, enhancing the model’s ability to capture both transient and steady-state features.

\subsection{Evaluation Setup}
\label{sec: study1 met}
\textbf{Setup.} Our system was developed under Python 3.8 and PyTorch 2.1 framework, tested for performance on an NVIDIA GeForce RTX 4090 GPU.  All random seeds were set to 2024.

\textbf{Train and test split.} In this subject-independent experiment, a five-fold cross-validation approach was used. The dataset was divided such that each fold’s test set contained 20\% of the subjects, while the remaining 80\% of subjects were used for training and validation. For each fold, 1,000 static pairs and 1,000 motion pairs were randomly selected, with each set consisting of 500 positive pairs and 500 negative pairs. These pairs were evenly sampled from the respective subjects (e.g., 5 subjects × 200 samples = 1,000 pairs). Positive pairs were generated by randomly selecting segments from the same subject, while negative pairs were created by pairing segments from different subjects. This method ensures a more balanced evaluation of performance across varying data quality and subject independence.

\textbf{Evaluation metrics.} We used several metrics to evaluate performance, including Area Under the Curve (AUC) and Equal Error Rate (EER), both of which provide threshold-independent insights. Higher AUC and lower EER values indicate better subject differentiation. Additionally, at a fixed False Rejection Rate (FRR) of 0.10, we measured the False Acceptance Rate (FAR) for each approach. Fixing the FRR allows us to simulate real-world conditions where a specific level of rejection tolerance is acceptable, ensuring a consistent benchmark for comparison. In practice, FAR is often a critical focus, as it indicates the likelihood of mistakenly accepting an unauthorized user, which is particularly important for security-sensitive applications. This fixed-FRR evaluation complements threshold-independent metrics, providing a more comprehensive understanding of the methods’ performance under realistic conditions. Reported metrics are averaged from 5 folds' results.













%% file: Sections/3-0-Feasible-Study.tex
\subsection{Feasibility Study with Open Finger PPG Datasets}

\label{sec: study 0}

\subsubsection{\textbf{Dataset}}

     \begin{table}[htp]
\centering
\caption{Detailed information on the datasets used in the experiment}
\label{tab:datasets}
\begin{tabularx}{\textwidth}{cccccc}
\toprule
Dataset Name & Subject Number & Sampling Rate & 
Collection Device & Position & Activity Label \\ 
\midrule
\multirow{1}{*}{MIMIC-III~\cite{johnson2016mimic}} & \textasciitilde 1400 & 125Hz & - & Fingertip & - \\
\addlinespace
BIDMC~\cite{7748483} & 53 & 125Hz & -  & Fingertip & - \\
\addlinespace
\multirow{1}{*}{MMPD-S~\cite{tang2023mmpd}} & 33 & 30Hz &  HKG-07C+ & Fingertip & \checkmark \\
\addlinespace
\multirow{1}{*}{DaLiA~\cite{misc_ppg-dalia_495}} & 15 & 64Hz & Empatica E4 & Wrist & \checkmark \\
\addlinespace
\midrule
\multirow{2}{*}{ANT-Motion} & 30 & 250Hz & Maxim & Wrist & \checkmark \\
 & 30 & 250Hz & Goodix & Wrist & \checkmark\\
 \multirow{1}{*}{ANT-Time} & 32 & 250Hz & Maxim & Wrist & \checkmark \\
\bottomrule
\end{tabularx}
\end{table}


Considering the vast diversity and widespread availability of well-established PPG datasets~\cite{johnson2016mimic,toye2023vital,7748483,tang2023mmpd,tang2023alpha,tang2024camerabased,tang2024spikingphysformer}, we first evaluated our proposed block architecture, RAPID, using those open-source datasets. In feasibility experiment, we utilized three open-source datasets: MIMIC-III~\cite{johnson2016mimic}, BIDMC~\cite{7748483}, and MMPD-S~\cite{tang2023mmpd}, as detailed in Table \ref{tab:datasets}. These datasets primarily feature single-channel green light PPG signals and are collected from fingertip sensors for medical use.

\subsubsection{\textbf{Baseline Algorithms}}

We constructed the RAPID model using two blocks and compared it with existing methods. InceptionTime~\cite{ismail2020inceptiontime}, known for time series processing, performs well but is too large to deploy on wearable devices. CorNET~\cite{cornet}, which combines LSTM and CNN architectures, has shown decent results in PPG-based heart rate estimation and identity verification.  CNN\_MFFD~\cite{wan2024deep} also demonstrates robust performance using a pure CNN design. We also adapted CNN\_LSTM~\cite{fotiadou2021cnnlstm}, originally used for fetal heart rate estimation, leveraging its effective frequency domain feature extraction capabilities.

\subsubsection{\textbf{RAPID Performs Best in Feasibility Study}}

\begin{table}[htp]
	\footnotesize
	\caption{Comparison of basic model performance on identity verification task}
	\label{tab:base model}
	\centering
	\setlength\tabcolsep{3.5pt} 
	\begin{tabular}{r|c|ccc|ccc|cccc}
        \toprule
		&  & \multicolumn{3}{c}{\textbf{MMPD-S}} &  \multicolumn{3}{c}{\textbf{MIMIC-III}}  & \multicolumn{3}{c}{\textbf{BIDMC}}   \\
        \textbf{Model} & PARAMS$\downarrow$ & AUC$\uparrow$ & EER$\downarrow$ & FAR$\downarrow$ & AUC$\uparrow$ & EER$\downarrow$ & FAR$\downarrow$ & AUC$\uparrow$ & EER$\downarrow$ & FAR$\downarrow$ &  \\ \hline \hline
        RAPID & \textbf{38752} & \textbf{0.95} & \textbf{0.13} & \textbf{0.15} & \textbf{0.97} & \textbf{0.07} & \textbf{0.03} & \textbf{0.97} & {0.08} & \textbf{0.05} \\ 
        InceptionTime \citep{ismail2020inceptiontime} & 471680 & {0.93} & {0.15}  & 0.22 & \textbf{0.97} & {0.08}  & {0.04} & \textbf{0.97} & \textbf{0.07} & \textbf{0.05} \\
        CorNET \citep{cornet} & {88896} & 0.91 & {0.18} & {0.29} & \textbf{0.97} & {0.12} & 0.16 & {0.94} & 0.10 & {0.11} \\
        CNN\_MFFD \citep{wan2024deep} & 99456 & 0.85  & 0.23  & 0.44 & 0.96  & {0.09}  & 0.07 & 0.91 & 0.16 & 0.26 \\
        CNN\_LSTM~\cite{fotiadou2021cnnlstm} & 571264 & 0.84 & 0.24 & 0.44 & 0.93 & 0.11 & 0.15 & 0.86 & 0.23 & 0.50 \\ 
        \bottomrule 
        
   \end{tabular}
      \\
   \tiny
   AUC = Area Under the Curve, EER = Equal Error Rate, FAR = False Accept Rate (When False Reject Rate = 0.10).
\end{table}

As outlined in Table \ref{tab:base model}, our RAPID model, demonstrates strong performance in all dataset tests, achieving AUC scores of 0.95 on the MMPD-S\cite{tang2023mmpd}, 0.97 on the MIMIC-III~\cite{johnson2016mimic}, and 0.97 on the BIDMC\cite{7748483} datasets, outperforming all competing baselines. The closest competitor, InceptionTime~\cite{ismail2020inceptiontime}, was substantially larger in model size, highlighting RAPID’s efficiency and capability in extracting meaningful PPG features.

%% file: Sections/3-1-Motion-Study.tex
\section{Reliable PPG Authentication on Wrist-worn Devices}
\subsection{Study 1: Reliable PPG Authentication on Daily Scenarios}
\label{sec: study 1}
To validate the performance of PPG authentication in real-life scenarios, we conducted a user study using wrist-based PPG sensors. The study evaluated the RAPID model without the quality assessment module on datasets containing both static and motion PPG signals, highlighting the challenges posed by daily activities and motion artifacts.  Subsequently, we introduced our MTL-RAPID model, enhanced with quality assessment module and identity verification module. This method demonstrated robust performance on our collected ANT dataset, effectively addressing the issues associated with poor-quality PPG signals in real-world applications.

\subsubsection{\textbf{Participants and Apparatus}}

\begin{figure*}[htp]
  \includegraphics[width=\textwidth]{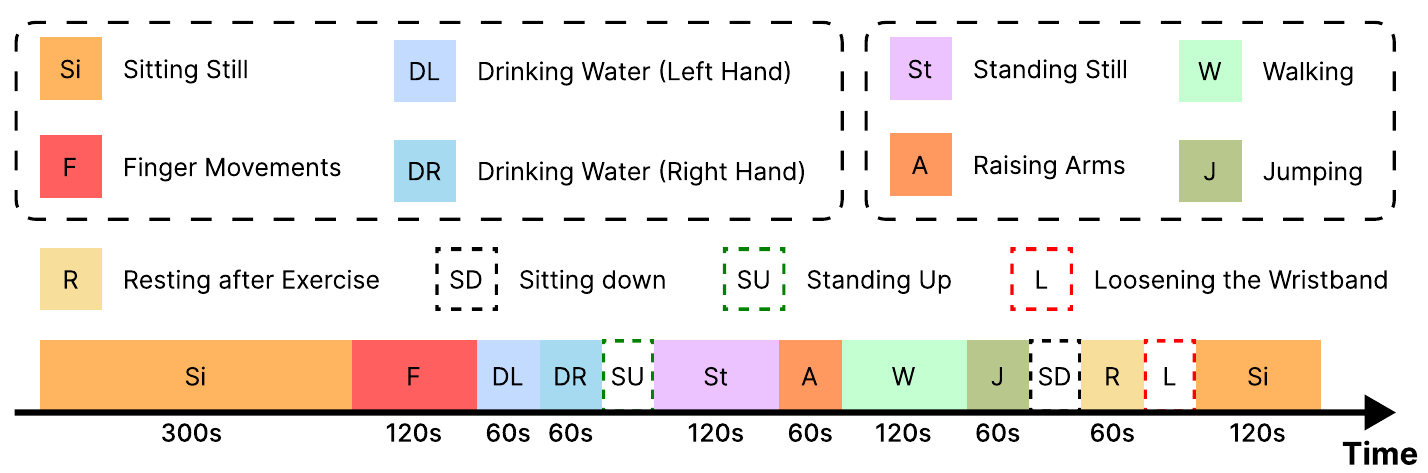}
  \vspace{-0.7cm}
  \caption{\textbf{Procedure for data collection.} ANT motion dataset fetures 10 activities (ranging from 60 to 300 seconds) with the re-worn process.}
  \label{fig:ANT dataset} 

\end{figure*}

We recruited 30 participants (two of them are not willing to disclose data and are not included in the following experiments) from our research institution. Among the 28 participants, all were right-handed, with an average age of 30.0 years (SD = 3.04). The group included 16 males (57.1\%) and 12 females (42.9\%). Skin tones, classified using the Monk Scale\footnote{\url{https://en.wikipedia.org/wiki/Monk_Skin_Tone_Scale}}, were primarily type 3 (60.7\%), followed by type 4 (21.4\%), type 2 (10.7\%), and type 5 (7.1\%). Finger conditions included normal (67.9\%), moist (14.3\%), dry (10.7\%), dirty (3.6\%), and peeling (3.6\%). This diverse sample provides a robust foundation for our study. The study protocols were reviewed and approved by the university's Institutional Review Board (IRB). Each participant wore wristbands equipped with two different PPG sensors: Goodix\footnote{\url{https://www.goodix.com/en/product/sensors/health_sensors/gh3220t}} on the left wrist and Maxim\footnote{\url{https://www.analog.com/en/resources/reference-designs/maxrefdes280.html}} on the right. After obtaining consent, participants were asked to perform 10 daily activities, including sitting, standing, walking, and jumping. These activities, detailed in Figure \ref{fig:ANT dataset}, were designed to simulate typical real-world usage without restricting specific movements, ensuring the data closely reflects daily behavior.

\begin{figure}[htbp]
    \centering
    \begin{subfigure}[b]{0.49\textwidth}
        \centering
        \includegraphics[width=0.6\textwidth, angle=270, keepaspectratio]{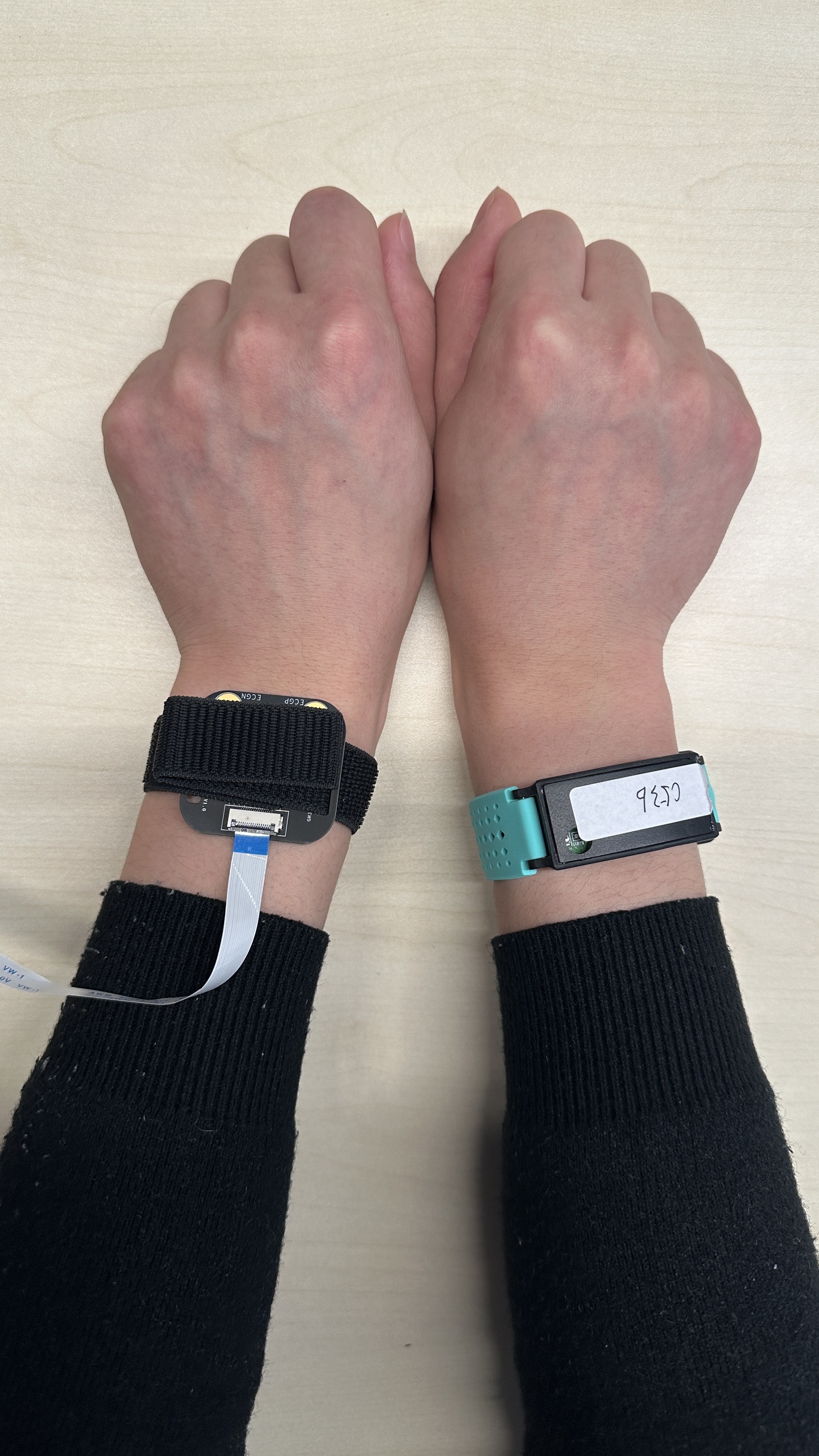} 
        \caption{Participants wearing wrist-based sensors.}
        \label{fig:subfig1}
    \end{subfigure}
    \hfill
    \begin{subfigure}[b]{0.49\textwidth}
        \centering
        \includegraphics[width=0.6\textwidth, angle=90, keepaspectratio]{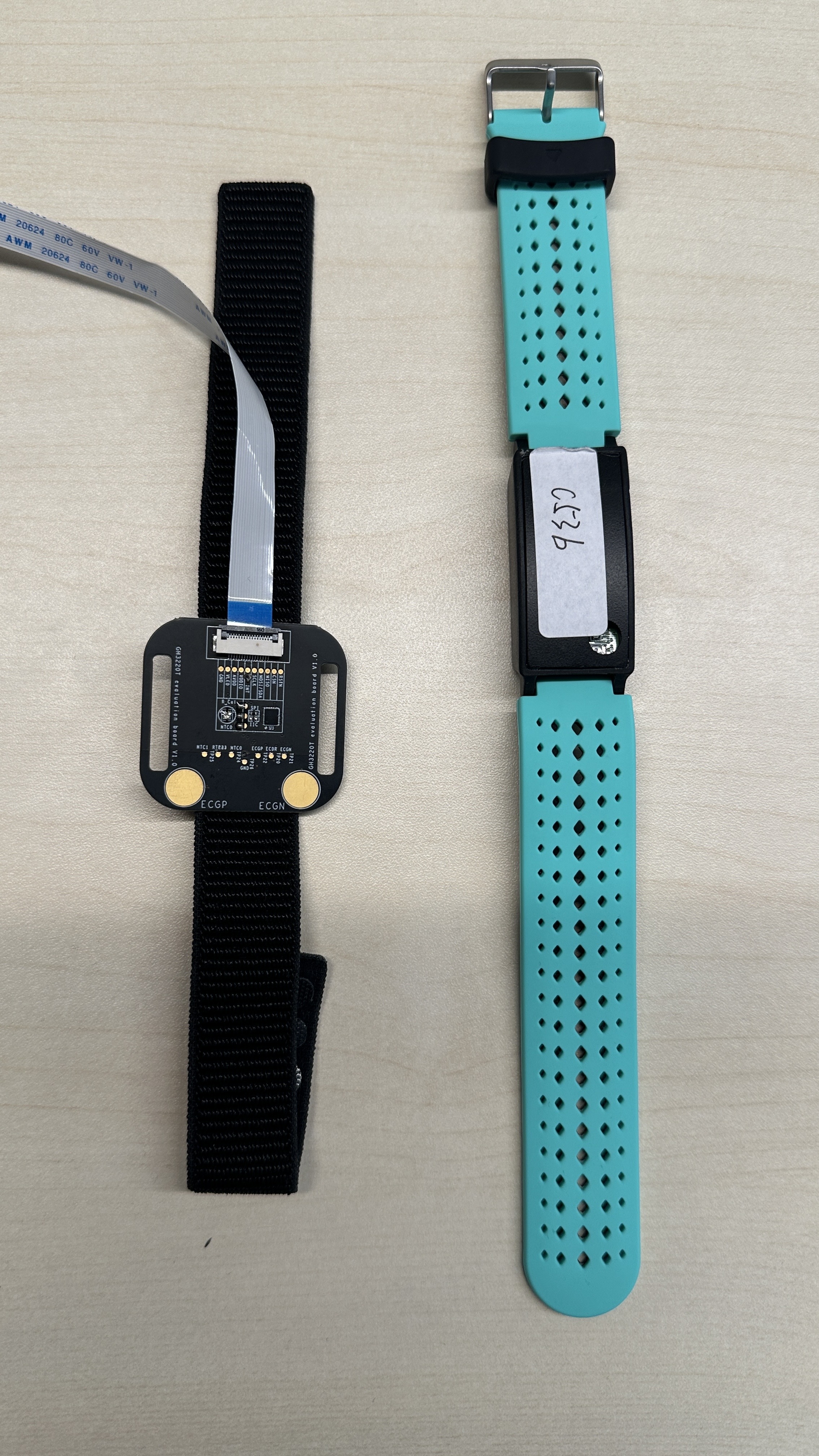} 
        \caption{Goodix and Maxim sensors.}
        \label{fig:subfig2}
    \end{subfigure}
    \caption{\textbf{Devices in the user study.}}
    \label{fig:mainfig}
\end{figure}

The data collected from these activities forms the Activity Noise Tracking (ANT) dataset, which addresses the limitations of prior datasets. Unlike other publicly available single-channel fingertip PPG datasets, such as MIMIC-III~\cite{johnson2016mimic} and BIDMC~\cite{7748483}, ANT includes three-channel data (red, green, and infrared) from wrist-worn sensors. This multi-channel approach offers a more comprehensive capture of user signals, as different wavelengths provide complementary information about the user’s physiological state. Additionally, wrist-worn sensors are more suitable for everyday use compared to fingertip sensors, making the ANT dataset more relevant for evaluating systems in realistic settings.

\subsubsection{\textbf{Experiments Setup}}

To demonstrate the necessity and reliability of the MTL-RAPID architecture, we conducted experiments on the motion data with the RAPID model and the MTL-RAPID model.


\textbf{Experiments on the impact of motion artifacts.} We evaluated the performance of five models—InceptionTime \citep{ismail2020inceptiontime}, CorNET \citep{cornet}, CNN\_MFFD \citep{wan2024deep}, CNN\_LSTM \citep{fotiadou2021cnnlstm}, and our RAPID model using the ANT dataset collected from Maxim and Goodix sensors, along with the DaLiA~\cite{misc_ppg-dalia_495} dataset. In this experiment as Table \ref{tab: motion data}, we first trained all models exclusively on static data to establish baseline performance. We then compared the performance on the complete datasets (static and motion data, denoted as mixed data) to assess how motion noise impacted identity verification accuracy.
\label{sec: life scenario}

\textbf{Experiments on MTL-RAPID effectiveness.} To assess the reliability of MTL-RAPID in real-world scenarios, we designed two distinct experimental strategies. Our MTL-RAPID model was trained jointly on diverse activity data to integrate signal quality assessment with identity verification. In comparison, baseline methods (InceptionTime~\cite{ismail2020inceptiontime}, CorNET~\cite{cornet}, CNN\_MFFD~\cite{wan2024deep}, CNN\_LSTM~\cite{fotiadou2021cnnlstm}) used separate training phases: their quality filters were trained on mixed data with activity labels, while verification modules were trained only on stable sitting data.

The evaluation process mirrored real-world usage through a two-stage filtering system. For each test set (2,000 sample pairs), signals were first screened by quality assessment modules with only qualified samples progressing to identity verification. This approach automatically excludes unreliable PPG segments before authentication attempts occur, closely replicating actual smartwatch operating conditions.

\subsubsection{\textbf{Results}}

\textbf{Baselines and RAPID method fail on datasets with motion artifacts.} As shown in Table \ref{tab: motion data}, the RAPID model outperformed baseline methods in most cases, achieving an AUC of 0.98 on the ANT\_Maxim dataset while maintaining the smallest model size.

On mixed datasets, RAPID maintained superior performance due to its noise robustness, while all models declined in performance, reflecting the challenges of real-world applications with motion artifacts. This suggests that in practical scenarios, relying solely on identity verification models may not produce reliable results without addressing signal quality.

Experiments show that PPG signals fluctuate significantly in real-life scenarios due to sensor placement and heart rate variations. These fluctuations significantly degrade the reliability of identity verification. For example, during activities like running or gym workouts, motion noise and temporal changes often lead to incorrect identification.

To address this, we designed the MTL-RAPID model. The model evaluates data quality before proceeding with identity verification, ensuring verification is only conducted when signal quality is adequate. This approach improves system reliability and adaptability in dynamic environments.

\begin{table}[htp]
	\footnotesize
	\caption{The Impact of Motion Data on Identity Verification Performance}
	\label{tab: motion data}
	\centering
	\setlength\tabcolsep{3.5pt} 
	\begin{tabular}{r|c|ccc|ccc|cccc}
        \toprule
		&  & \multicolumn{3}{c}{\textbf{ANT\_Maxim\_static}} &  \multicolumn{3}{c}{\textbf{DaLiA\_static}}  & \multicolumn{3}{c}{\textbf{ANT\_Goodix\_static}}   \\
        \textbf{Model} & PARAMS$\downarrow$ & AUC$\uparrow$ & EER$\downarrow$ & FAR$\downarrow$ & AUC$\uparrow$ & EER$\downarrow$ & FAR$\downarrow$ & AUC$\uparrow$ & EER$\downarrow$ & FAR$\downarrow$ &  \\ \hline \hline
        RAPID & \textbf{38752} & \textbf{0.98} & \textbf{0.06} & \textbf{0.03} & 0.85 & 0.23 & 0.51 & \textbf{0.97} & \textbf{0.09} & \textbf{0.09}  \\ 
        InceptionTime \citep{ismail2020inceptiontime} & 471680 & 0.97 & 0.09 & 0.08 & \textbf{0.87} & \textbf{0.20} & \textbf{0.36 } & 0.95 & 0.12  & 0.17 \\
        CorNET \citep{cornet} & {88896} & 0.95 & 0.11 & 0.12 & 0.81 & 0.25 & 0.61 &  0.91 & 0.17 & 0.28  \\
        CNN\_MFFD \citep{wan2024deep} & 99456 & 0.89  & 0.19  & 0.33 & 0.84 & 0.24 & 0.47 & 0.75 & 0.31  & 0.67  \\
        CNN\_LSTM~\cite{fotiadou2021cnnlstm} & 571264 & 0.80 & 0.27 & 0.54 & 0.75 & 0.32 & 0.66 & 0.66 & 0.39 & 0.76 \\ 
        \bottomrule 
        \toprule
		&  & \multicolumn{3}{c}{\textbf{ANT\_Maxim\_mixed}} &  \multicolumn{3}{c}{\textbf{DaLiA\_mixed}}  & \multicolumn{3}{c}{\textbf{ANT\_Goodix\_mixed}}   \\
        \textbf{Model} & PARAMS$\downarrow$ & AUC$\uparrow$ & EER$\downarrow$ & FAR$\downarrow$ & AUC$\uparrow$ & EER$\downarrow$ & FAR$\downarrow$ & AUC$\uparrow$ & EER$\downarrow$ & FAR$\downarrow$ &  \\ \hline \hline
        RAPID & \textbf{38752} & \textbf{0.82} & \textbf{0.26} & \textbf{0.55} & 0.73 & 0.34 & 0.76 & \textbf{0.79} & \textbf{0.29} & \textbf{0.64}  \\ 
        InceptionTime \citep{ismail2020inceptiontime} & 471680 & 0.81 & 0.27 & 0.63 & \textbf{0.75} & \textbf{0.30} & \textbf{0.69}  & 0.76 & 0.30  & 0.66 \\
        CorNET \citep{cornet} & {88896} & 0.78 & 0.29 & 0.64 & 0.70 & 0.34 & 0.83 & 0.74 & 0.32 & 0.72  \\
        CNN\_MFFD \citep{wan2024deep} & 99456 & 0.73  & 0.33  & 0.72 & 0.74 & 0.32 & 0.66 & 0.66 & 0.39  & 0.80  \\
        CNN\_LSTM~\cite{fotiadou2021cnnlstm} & 571264 & 0.68 & 0.37 & 0.75 & 0.59 & 0.43 & 0.87 & 0.59 & 0.44 & 0.85 \\ 
        \bottomrule 
   \end{tabular}
      \\
   \tiny
   AUC = Area Under the Curve, EER = Equal Error Rate, FAR = False Accept Rate (When False Reject Rate = 0.10).

\end{table}

\begin{figure*}[htp]
\centering
  \includegraphics[width=\textwidth]{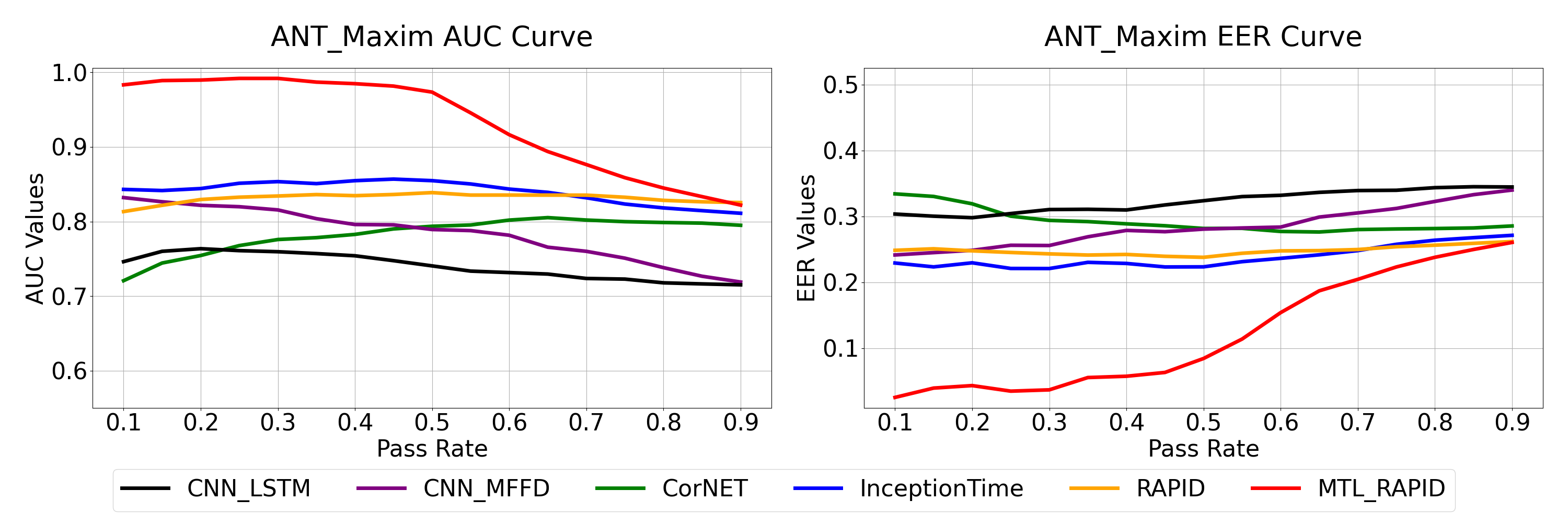}
    \vspace{-0.3cm}
  \caption{\textbf{The results of cross-activity experiments on ANT\_Maxim dataset.} MTL-RAPID got the best AUC of \textbf{99.2}\% and an EER of \textbf{3.5}\%, outperforming all baselines. The x-axis represents the pass rate after filtering, and the y-axis displays the AUC and EER of the identity verification task.}
  \label{fig: MTL Maxim}
\end{figure*}

\begin{figure*}[htp]
\centering
  \includegraphics[width=\textwidth]{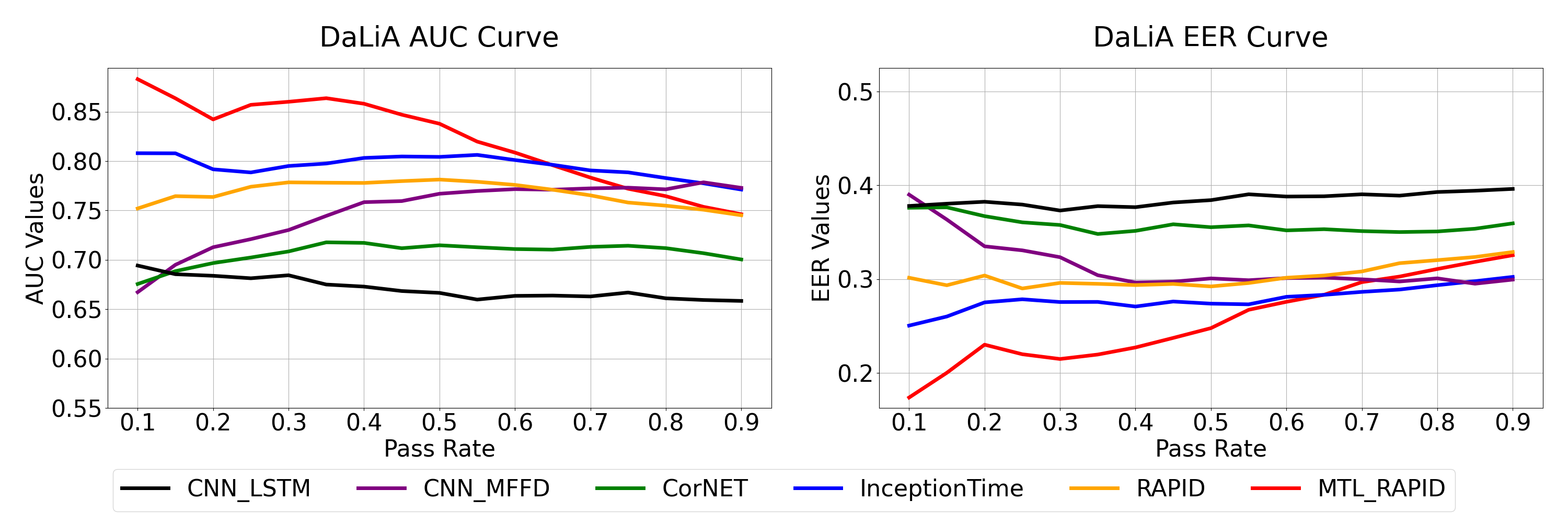}
      \vspace{-0.3cm}
  \caption{\textbf{The results of cross-activity experiments on DaLiA dataset.} MTL-RAPID got the best AUC of \textbf{88.3}\% and an EER of \textbf{17.4}\%, outperforming all baselines. The x-axis represents the pass rate after filtering, and the y-axis displays the AUC and EER of the identity verification task.}
  \label{fig: MTL DaLiA}
\end{figure*}

\begin{figure*}[htp]
\centering
  \includegraphics[width=\textwidth]{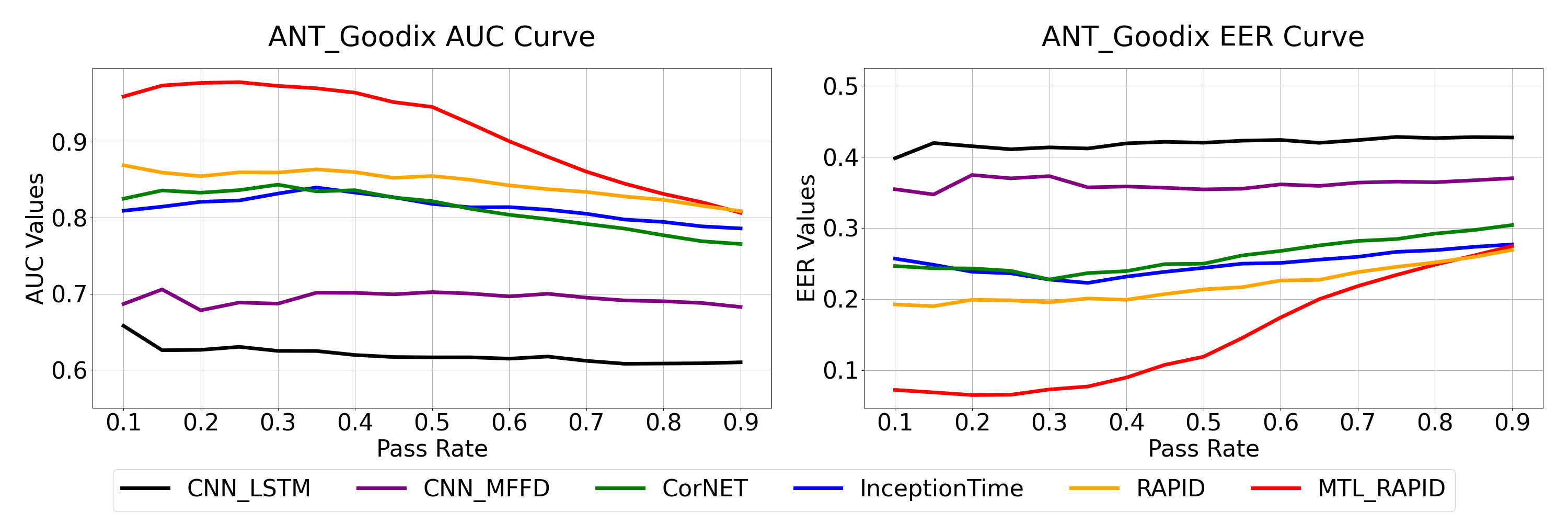}
      \vspace{-0.3cm}
  \caption{\textbf{The results of cross-activity experiments on ANT\_Goodix dataset.} MTL-RAPID got the best AUC of \textbf{97.8}\% and an EER of \textbf{6.5}\%, outperforming all baselines. The x-axis represents the pass rate after filtering, and the y-axis displays the AUC and EER of the identity verification task.}
  \label{fig: MTL Goodix}
\end{figure*}

\textbf{MTL-RAPID outperforms sequentially connected models.} We compared MTL-RAPID (Section \ref{sec: MTL model}) with traditional approaches using separate quality classifiers and identity verification models. By adjusting the final classification threshold, we ensured all models were validated on an equal number of samples. The performances of ANT\_Maxim, DaLiA~\cite{misc_ppg-dalia_495} and ANT\_Goodix are visualized in Figure \ref{fig: MTL Maxim}, Figure \ref{fig: MTL DaLiA} and Figure \ref{fig: MTL Goodix}, where the x-axis represents the pass rate after filtering, and the y-axis displays the AUC and EER of the identity verification task.

Excluding noisy samples improves AUC and reduces EER for identity verification. Notably, MTL-RAPID consistently outperforms separate RAPID models and other methods, regardless of the number of remaining sample pairs. This advantage stems from the simultaneous training of the quality assessment and authentication modules in MTL-RAPID, enabling the model to learn inter relationships between these tasks. At a pass rate of approximately 25\%, MTL-RAPID achieved an AUC of \textbf{99.2}\% and an EER of \textbf{3.5}\% on the ANT\_Maxim dataset, demonstrating its reliability in handling noisy data from daily activities.

Adjusting classification thresholds enables flexible trade-offs between data availability and authentication performance. For instance, in low-risk scenarios like message checks, a higher pass rate enhances convenience. Conversely, in high-security applications like payment authentication, stricter filtering prioritizes accuracy and security, ensuring only high-quality data is used for authentication.



%% file: Sections/4-LongTerm-Study.tex
\subsection{Study 2: Reliable PPG Authentication on Cross-Day Scenarios}
\label{sec: study 2}

Considering the daily wearing patterns of smartwatches, developing a robust cross-day authentication model is crucial. To address this, we conducted a user study focusing primarily on the performance and stability of the system in cross-time verification scenarios. This study aimed to evaluate the models' ability to maintain reliable authentication across different days, highlighting the challenges in long-term usability.

\subsubsection{\textbf{Participants and Apparatus}}



To evaluate the long-term reliability of PPG-based authentication, we conducted a follow-up user study four months after the movement study. In this study, we recruited 32 participants and collected data over five consecutive days. Among them, 28 were right-handed (87.5\%) and 4 were left-handed (12.5\%). The average age was 30.97 years (SD = 3.55). The group comprised 21 males (65.6\%) and 11 females (34.4\%). Skin tones were distributed as type 3 (31.3\%), type 2 (25.0\%), type 4 (21.9\%), and type 5 (12.5\%). Regarding finger conditions, 28 participants (87.5\%) had normal fingers, while 3 (9.4\%) had moist fingers, and 1 participant (3.1\%) had peeling.

After obtaining consent, participants were fitted with wristbands equipped with Maxim\footnote{\url{https://www.analog.com/en/resources/reference-designs/maxrefdes280.html}} sensors. They were instructed to remain seated or office status(allowing slight hand and body movements while seated). We collected 5 minutes of three-channel PPG signals (red, green, and infrared) for each posture. Over the following four days, participants returned to collect additional data under the same conditions.

\subsubsection{\textbf{Experiments on Cross-day Authentication Scenario.}}
\label{sec: cross-day}
To assess the accuracy of identity verification over different time intervals following user registration, we tested the MTL-RAPID and other models at intervals of 1, 2, 3, and 4 days. This evaluation aimed to validate the model's robustness to physiological variations over time.

We designed a two-phase training strategy to evaluate authentication performance across different days. First, we trained the base MTL-RAPID model using data from 16 subjects in the ANT motion dataset, as described in Section \ref{sec: study 1}. Subsequently, we fine-tuned the model using static sitting data from the training set corresponding to the test days, adapting it to temporal variations. The training and test sets were split using a five-fold subject-independent approach, as described in Section \ref{sec: study1 met}.

\subsubsection{\textbf{RAPID Series Outperform Baselines on Cross-Day Scenarios}}

\begin{figure*}[htp]
\centering
  \includegraphics[width=\textwidth]{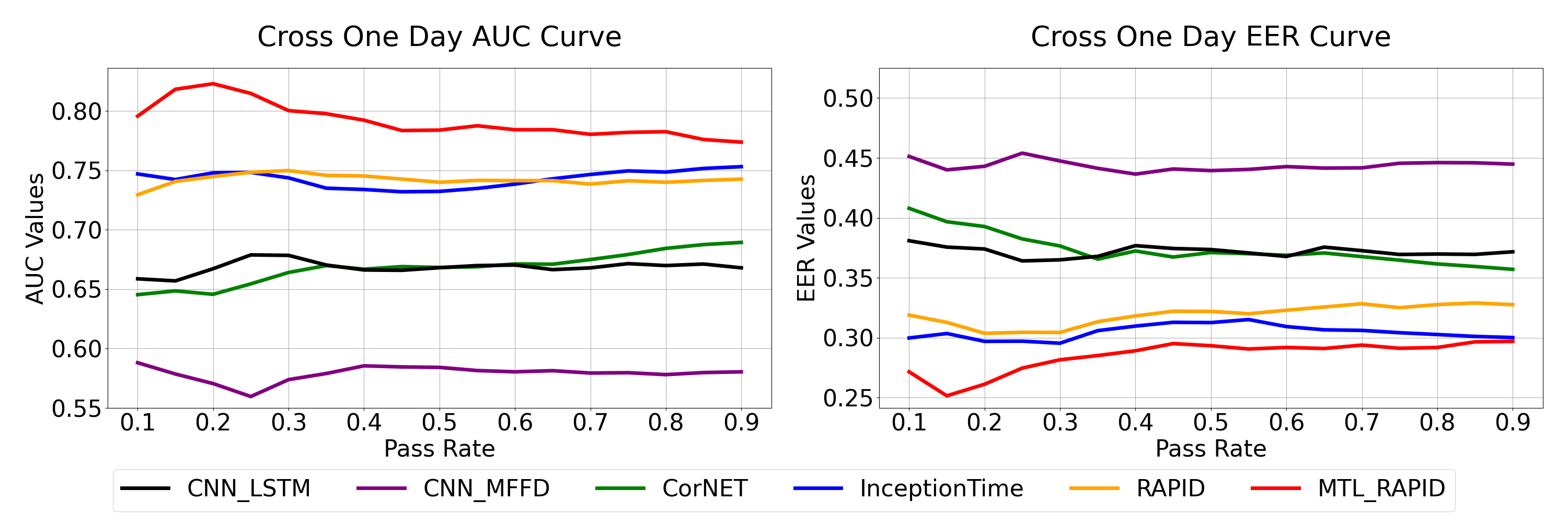}
      \vspace{-0.3cm}
  \caption{\textbf{The results of cross one day experiments on ANT\_Maxim dataset.} MTL-RAPID got the best AUC of \textbf{82.3}\% and an EER of \textbf{26.1}\%, outperforming all baselines. The x-axis represents the pass rate after filtering, and the y-axis displays the AUC and EER of the identity verification task.}
  \label{fig: MTL 1 day}
\end{figure*}

\begin{figure*}[htp]
\centering
  \includegraphics[width=\textwidth]{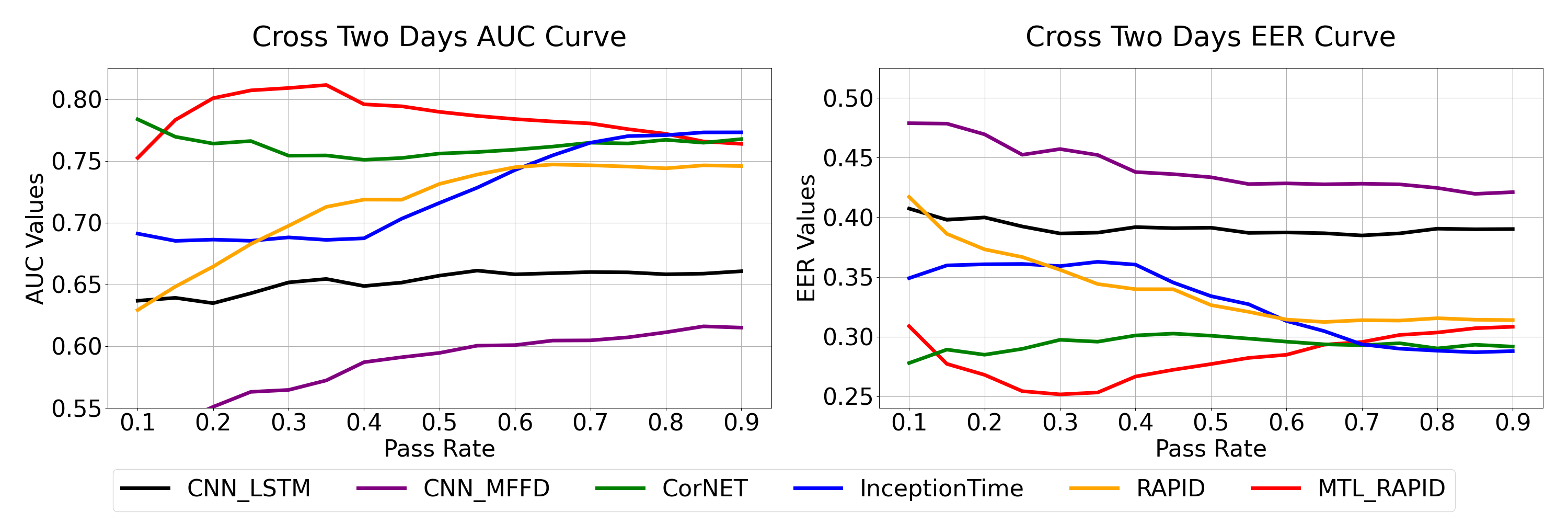}
      \vspace{-0.3cm}
  \caption{\textbf{The results of cross two days experiments on ANT\_Maxim dataset.} MTL-RAPID got the best AUC of \textbf{81.4}\% and an EER of \textbf{5.3}\%, outperforming all baselines. The x-axis represents the pass rate after filtering, and the y-axis displays the AUC and EER of the identity verification task.}
  \label{fig: MTL 2 day}
\end{figure*}

\begin{figure*}[htp]
\centering
  \includegraphics[width=\textwidth]{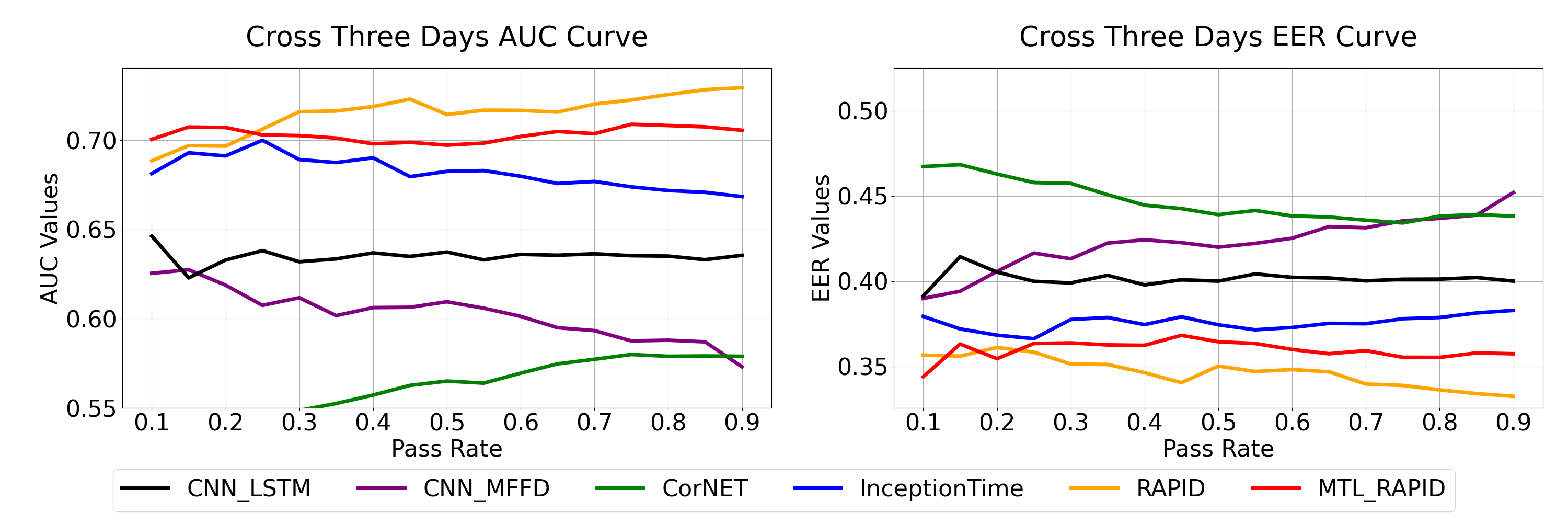}
      \vspace{-0.3cm}
  \caption{\textbf{The results of cross three days experiments on ANT\_Maxim dataset.} RAPID got the best AUC of \textbf{72.9}\% and an EER of \textbf{33.2}\%, outperforming all baselines. The x-axis represents the pass rate after filtering, and the y-axis displays the AUC and EER of the identity verification task.}
  \label{fig: MTL 3 day}
\end{figure*}

\begin{figure*}[htp]
\centering
  \includegraphics[width=\textwidth]{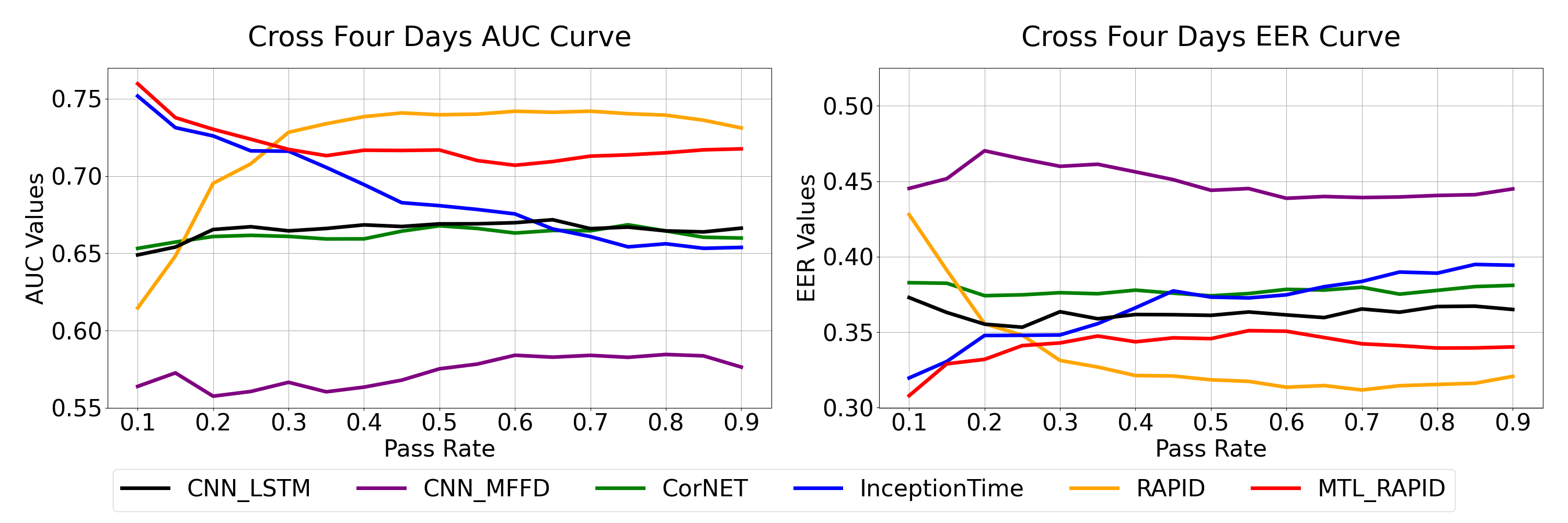}
      \vspace{-0.3cm}
  \caption{\textbf{The results of cross four days experiments on ANT\_Maxim dataset.} MTL-RAPID got the best AUC of \textbf{76.0}\% and an EER of \textbf{30.8}\%, outperforming all baselines. The x-axis represents the pass rate after filtering, and the y-axis displays the AUC and EER of the identity verification task.}
  \label{fig: MTL 4 day}
\end{figure*}

Our experimental results demonstrate RAPID series got the best performance in cross-day authentication scenarios, while also revealing the inherent challenges of long-term biometric verification. As shown in Figures \ref{fig: MTL 1 day}-\ref{fig: MTL 4 day}, the RAPID series consistently outperformed all baselines across all tested intervals. In the one-day scenario (Figure \ref{fig: MTL 1 day}), MTL-RAPID achieved an AUC of 82.3\% and EER of 26.1\%, surpassing InceptionTime by 7\% (AUC: 82.3\% vs. 75.3\%). This lead was maintained in the two-day scenario (Figure \ref{fig: MTL 2 day}) with an AUC of 81.4\% and EER of 5.3\%, outperforming CorNet (AUC: 81.4\% vs. 78.4\%). While all methods exhibited performance degradation compared to the cross-activity dataset, the RAPID series showed a more stable trend. On the third day, the original RAPID model achieved the best performance with an AUC of 72.9\% and EER of 33.2\%, slightly outperforming MTL-RAPID (AUC: 70.9\%). 

The RAPID series' consistent outperformance highlights the effectiveness of its design, with MTL-RAPID excelling in short to medium time frames and the original RAPID model demonstrating strong performance in specific long-term scenarios. Despite the challenges posed by physiological variations, these results underscore the potential of the RAPID series for real-world applications and emphasize the need for future research to address the fundamental limitations of long-term biometric verification.


%% file: Sections/6-Ablation-Study.tex
\subsection{Ablation Study}
The ablation study results highlight the importance of selecting optimal parameters for PPG-based identity authentication. By systematically evaluating the impact of window length, sampling rate, and multi-channel signals, we identified configurations that balance performance and practicality. These findings not only validate the robustness of the RAPID model but also provide insights into the trade-offs between accuracy and real-world feasibility.
\subsubsection{\textbf{Window Length Selection}}\label{sec:length}

When preprocessing signals, segmenting the PPG signals into equal-length PPG fragments is necessary. Different segment lengths can affect the model's performance in identity authentication tasks and influence feasibility in real-world applications. We trained and tested the RAPID model on PPG signals of varying lengths using five datasets: BIDMC\cite{7748483}, ANT\_Maxim, ANT\_Goodix, DaLiA\cite{misc_ppg-dalia_495}, and MMPD\cite{tang2023mmpd}. The results are recorded in Table \ref{tab:time_length}. For the BIDMC\cite{7748483} dataset, which has a sufficient amount of data per subject, we experimented with segment lengths of 25 seconds and 30 seconds. However, due to insufficient data per subject in the other four datasets, these longer segment lengths were tested only on the BIDMC\cite{7748483} dataset.

\begin{table}[H]
	\footnotesize
	\caption{The impact of PPG signal duration on identity authentication performance}
	\label{tab:time_length}
	\centering
	\setlength\tabcolsep{5pt} 
	\begin{tabular}{r|cc|cc|cc|cc|ccc}
	\toprule
		&\multicolumn{2}{c}{\textbf{BIDMC}} & \multicolumn{2}{c}{\textbf{ANT\_Maxim}} &  \multicolumn{2}{c}{\textbf{ANT\_Goodix}} &  \multicolumn{2}{c}{\textbf{DaLiA}} &  \multicolumn{2}{c}{\textbf{MMPD}}   \\
        \textbf{PPG Length} & AUC$\uparrow$ & EER$\downarrow$ & AUC$\uparrow$ & EER$\downarrow$ & AUC$\uparrow$ & EER$\downarrow$ & AUC$\uparrow$ & EER$\downarrow$ & AUC$\uparrow$ & EER$\downarrow$ &  \\ \hline \hline
        3s & {0.96} & 0.10 & {\textbf{0.97}} & {0.08} & 0.96 & 0.10 & \textbf{0.84} & \textbf{0.24} & 0.93 & 0.14 \\ 
        6s & 0.97 & 0.09 & \textbf{0.97} & \textbf{0.07} & 0.95 & 0.11 & 0.82 & 0.26 & 0.93 & 0.14 \\
        10s & {\textbf{0.98}} & \textbf{0.07} & {\textbf{0.97}} & {0.08} & \textbf{0.96} & \textbf{0.10} & 0.83 & 0.25 & \textbf{0.94} & \textbf{0.12} \\
        15s & 0.97 & \textbf{0.07} & 0.95 & 0.10 & 0.94 & 0.13 & 0.82 & 0.26 & 0.91 & 0.16 \\
        20s & 0.96 & 0.08 & 0.95 & 0.12 & 0.92 & 0.15 & 0.79 & 0.27 & 0.91 & 0.17 \\ 
        25s & 0.97 & 0.08 & \multicolumn{8}{c}{|} \\ 
        30s & 0.96 & 0.09 & \multicolumn{8}{c}{Not Applicable} \\ 
        \bottomrule
   \end{tabular}
   \\
   \tiny
   AUC = Area Under the Curve, EER = Equal Error Rate
\end{table}

As shown in Table \ref{tab:time_length}, the 6-second and 10-second segments achieve comparable performance across datasets (e.g., BIDMC: AUC=0.97 vs. 0.98; ANT\_Maxim: AUC=0.97 for both).   However, shorter segments (3s) exhibit degraded performance on motion-prone datasets like BIDMC (AUC=0.96 vs. 0.97 for 6s), while longer segments ($\geq$15s) reduce practicality without performance gains (MMPD AUC drops to 0.91 at 15s). This balance between robustness and usability justifies the 6-second choice.

\subsubsection{\textbf{Sampling Rate Selection}}\label{sec:fs}

To determine the optimal PPG sampling frequency, we conducted experiments using the RAPID model on four datasets with high original sampling frequencies: ANT\_Maxim, ANT\_Goodix, MIMIC-III\cite{johnson2016mimic}, and BIDMC\cite{7748483}. We tested the identity authentication performance at four commonly used sampling frequencies (30Hz, 60Hz, 125Hz, and 250Hz). The results are recorded in Table \ref{tab:sampling rate}. Since the original sampling frequencies of MIMIC-III\cite{johnson2016mimic} and BIDMC\cite{7748483} are only 125Hz, we did not perform the 250Hz tests on these two datasets.

\begin{table}[H]
	\footnotesize
	\caption{The impact of PPG signal's sampling rate on identity authentication performance}
	\label{tab:sampling rate}
	\centering
	\setlength\tabcolsep{6pt} 
	\begin{tabular}{r|cc|cc|cc|ccc}
	\toprule
		&\multicolumn{2}{c}{\textbf{ANT\_Maxim}} & \multicolumn{2}{c}{\textbf{ANT\_Goodix}} &  \multicolumn{2}{c}{\textbf{MIMIC-III}} &  \multicolumn{2}{c}{\textbf{BIDMC}} \\
        \textbf{Sampling Rate} & AUC$\uparrow$ & EER$\downarrow$ & AUC$\uparrow$ & EER$\downarrow$ & AUC$\uparrow$ & EER$\downarrow$ & AUC$\uparrow$ & EER$\downarrow$  \\ \hline \hline
        30Hz & 0.81 & 0.26 & 0.79 & 0.28 & 0.96 & 0.09 & \textbf{0.97} & \textbf{0.08}  \\ 
        60Hz & \textbf{0.97} & \textbf{0.08} & \textbf{0.95} & \textbf{0.11} & \textbf{0.97} & \textbf{0.08} & \textbf{0.97} & \textbf{0.08} \\
        125Hz & \textbf{0.98} & 0.08 & 0.95 & 0.12 & 0.96 & 0.08 & 0.96 & 0.08 \\
        250Hz & \textbf{0.98} & 0.07 & 0.96 & 0.10 & \multicolumn{4}{c}{Not Applicable} \\
        \bottomrule
   \end{tabular}
   \\
   \tiny
   AUC = Area Under the Curve, EER = Equal Error Rate
\end{table}

While 125Hz and 250Hz marginally improve ANT\_Maxim performance (AUC=0.98 vs. 0.97 at 60Hz), 60Hz achieves optimal or near-optimal results across all datasets (e.g., ANT\_Goodix AUC=0.95 vs. 0.96 at 250Hz) while halving computational requirements compared to higher rates. This makes 60Hz a pragmatic compromise between signal fidelity and efficiency.

\subsubsection{\textbf{Multi-Channels PPG}}\label{sec:multi-channels}

Due to the varying absorption rates of different wavelengths of light by the human body, PPG signals recorded under different wavelengths capture different information. We compared the training performance of single-channel green light and three-channel PPG on two ANT datasets to assess whether multi-channel signals provide better authentication results. The results are shown in Table \ref{tab:multi-channels}.

\begin{table}[H]
	\footnotesize
	\caption{The impact of the number of channels in PPG signals on identity authentication performance}
	\label{tab:multi-channels}
	\centering
	\setlength\tabcolsep{2pt} 
	\begin{tabular}{r|ccc|ccc}
	\toprule
		&\multicolumn{3}{c}{\textbf{ANT\_Maxim}} & \multicolumn{3}{c}{\textbf{ANT\_Goodix}} \\
        \textbf{Channels} & AUC$\uparrow$ & EER$\downarrow$ & FAR$\downarrow$ & AUC$\uparrow$ & EER$\downarrow$ & FAR$\downarrow$  \\ \hline \hline
        1 Channel & \textbf{0.97 $\pm$ 0.01} & \textbf{0.08 $\pm$ 0.02} & 0.07 $\pm$ 0.04  &  0.86 $\pm$ 0.06 & 0.21 $\pm$ 0.06 & 0.47 $\pm$ 0.22 \\ 
        3 Channels & \textbf{0.97 $\pm$ 0.01} & \textbf{0.08 $\pm$ 0.02} & \textbf{0.05 $\pm$ 0.03} & \textbf{0.95 $\pm$ 0.01} & \textbf{0.11 $\pm$ 0.01} & \textbf{0.13 $\pm$ 0.04} \\
        \bottomrule
   \end{tabular}
   \\
   \tiny
   AUC = Area Under the Curve, EER = Equal Error Rate, FAR = False Accept Rate (When False Reject Rate = 0.10).
\end{table}

Three-channel PPG demonstrates critical advantages: it maintains ANT\_Maxim's performance (AUC=0.97) while dramatically improving ANT\_Goodix's AUC from 0.86 to 0.95 and reducing FAR by 72\% (0.47 to 0.13). This spectral diversity captures complementary biometric features, particularly beneficial for devices with lower baseline performance.

Our ablation studies collectively justify the parameter choices: 6-second segments balance performance (BIDMC AUC=0.97) and usability; 60Hz sampling optimizes computational efficiency without sacrificing accuracy (ANT\_Maxim AUC=0.97 vs. 0.98 at 125Hz); and 3-channel PPG enhances reliability, especially for challenging scenarios (ANT\_Goodix FAR=0.13 vs. 0.47). These selections establish a robust foundation for practical PPG-based authentication systems.

%% file: Sections/5-Usable-Study.tex
\section{Usability Study}
\label{sec: study 3}

While we validated reliability in real-world, long-term scenarios, we also recognized usability as a crucial component influencing user preference.
We conducted a user study to evaluate the proposed PPG-based authentication method in comparison with PIN-based authentication, which is currently the most widely used commercially implemented authentication method for smartwatches. We aimed to evaluate the time duration required for registration and authentication, the success rates for both legitimate users and potential attackers, and the overall user experience. Feedback from participants and their test results were collected for analysis. The study protocols were reviewed and approved by the university's IRB. All participants were informed about the protocols and agreed to participate in the study.

\subsection{Participants and Apparatus}

\subsubsection{\textbf{Demographics}}

We recruited 16 participants from the students and faculties at the university. Among the participants, 9 were female, and 7 were male, with an average age of 23.6 years (SD = 2.76). All participants were right-handed and instructed to wear the smartwatch on their left wrist. Regarding smartwatch usage habits, 31.25\% of participants reported wearing a smartwatch daily, while 56.25\% stated that they seldom or never used smartwatches.

In terms of prior experience with smartwatch authentication, 58.3\% of participants had never used any authentication method on a smartwatch, 16.7\% had used PIN-based methods, and the remainder relied on mobile phones to activate their smartwatches. This latter approach was excluded from the study as it is not an independent authentication method and requires the use of additional devices. Furthermore, 75\% of participants agreed that implementing authentication methods is both meaningful and necessary for protecting privacy on smartwatches.

\subsubsection{\textbf{Devices}}\label{sec:usable study device}

\begin{figure}[htbp]
    \centering
    \begin{subfigure}[b]{0.49\textwidth}
        \centering
        \includegraphics[width=0.82\textwidth, keepaspectratio]{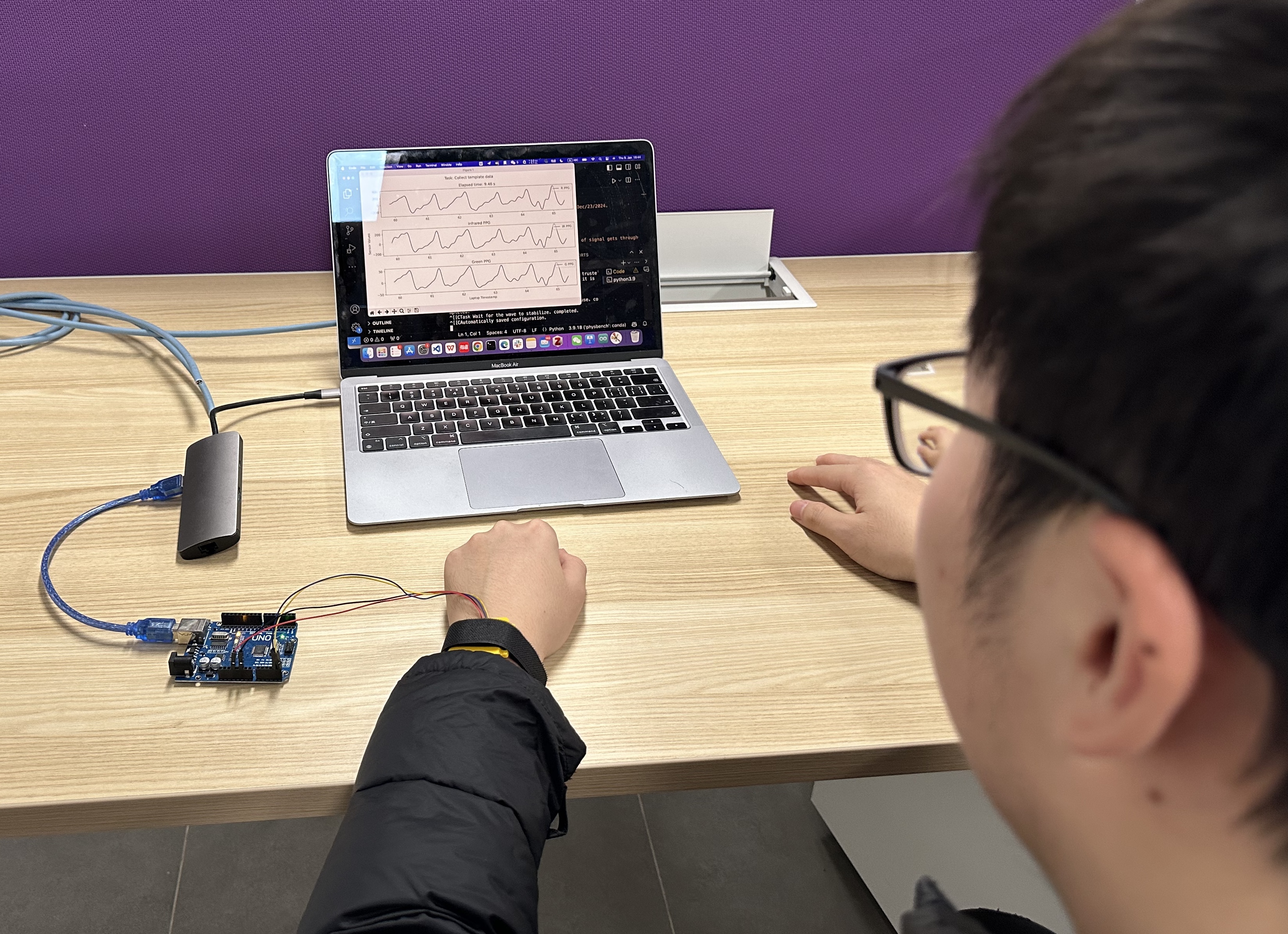} 
        \caption{Participants wearing the PPG watch.}
        \label{fig:usable participant}
    \end{subfigure}
    \hfill
    \begin{subfigure}[b]{0.49\textwidth}
        \centering
        \includegraphics[width=0.8\textwidth, keepaspectratio]{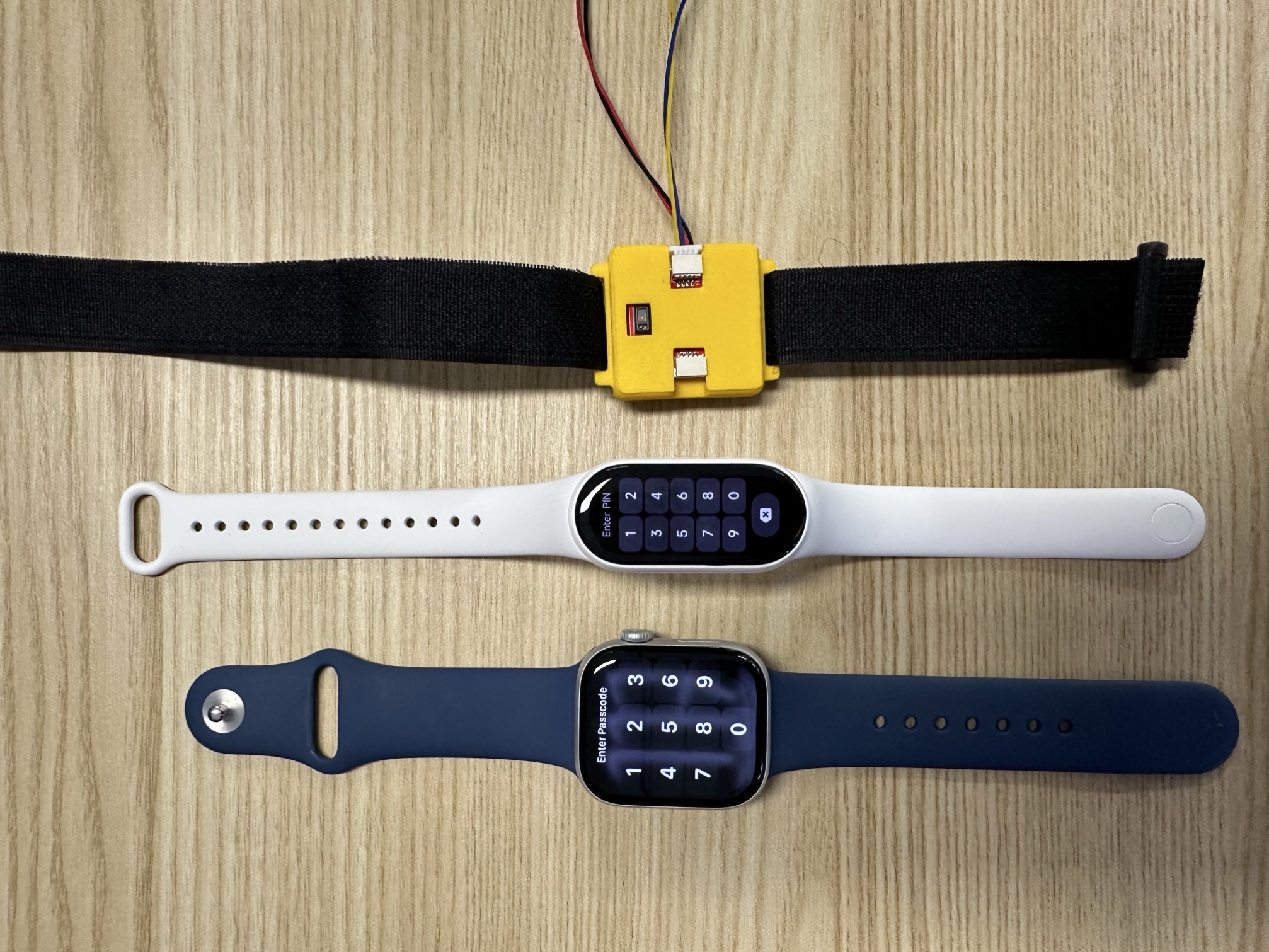} 
        \caption{PPG watch, Xiaomi Smart Band and Apple Watch.}
        \label{fig:subfig2}
    \end{subfigure}
    \caption{\textbf{Devices in the user study.}}
    \label{fig:usable devices}
\end{figure}

In the study, we utilized three smartwatches: a custom-built PPG watch designed for PPG-based authentication, and two commercially available smartwatches using PIN-based authentication, including an Apple Watch Series 10\footnote{\url{https://www.apple.com/apple-watch-series-10/}} and a Xiaomi Smart Band 9\footnote{\url{https://www.mi.com/global/product/xiaomi-smart-band-9/}}, as shown in Figure \ref{fig:usable devices}.

The custom PPG watch featured a MAX30101\footnote{\url{https://www.analog.com/en/products/max30101.html}} PPG sensor, a 3D-printed casing, and a strap. It was connected to an Arduino Uno board via USB, which interfaced with a MacBook Air and sampled at approximately 100 Hz. The collected signals were preprocessed using the method described in Section \ref{sec: preprocess}, then transmitted to a pretrained MTL-RAPID model, trained on the ANT\_Maxim dataset illustrated in Figure \ref{fig:ANT dataset}. The MTL-RAPID model ran on the CPU of the MacBook Air.

The Apple Watch, with a larger screen size ($1.81 \times 1.65$ inches, ~1.53 in$\hat{2}$), used a 4-digit PIN for authentication. While this setup provided easier access for users, it also increased the risk of successful attacks. In contrast, the Xiaomi Smart Band, with a smaller and narrower screen ($1.87 \times 0.43$ inches, ~0.80 in$\hat{2}$), employed a 6-digit PIN, offering stronger security but requiring more effort from users.

\subsection{Evaluation Setup}

\subsubsection{\textbf{Attack Model}}

In this study, we aim to compare the vulnerabilities of PIN-based and PPG-based authentication systems against unauthorized user attacks. To achieve this, we first describe the attack models for both methods.

Extensive prior research has examined attack models for PIN-based authentication, focusing primarily on the vulnerability of passwords to being observed or memorized by unauthorized users. Following this approach, we simulate attacks by allowing the experimenter to observe participants entering their PIN and then attempting to unlock the device using the observed PIN.

For PPG-based authentication, previous studies have primarily investigated attack models where remote PPG (rPPG) signals are extracted from facial videos to generate PPG data from other body regions, potentially compromising authentication systems \cite{hinatsu2022evaluation,liu2024summit}. However, these methods remain theoretical, requiring additional video data, and are not applicable to watch-only scenarios. In this experiment, we simulate attacks on PPG-based authentication by using a different user's wrist PPG signals to gain unauthorized access to the device.
\subsubsection{\textbf{Experiment Procedure}}

In this study, the experimenter first provided a brief introduction to the procedure and obtained informed consent from the participants.

During the registration process, the watch first collected PPG templates from the participants, which took at least 10 seconds. Once the templates were collected, the watch attempted authentication three times. Additionally, three random PPG signals, collected from different subjects prior to the experiment, were used to attempt an attack on the watch. A 15-second break was provided before each authentication attempt and attack. During template collection, the signals were sent to the MTL-RAPID model for quality assessment and were filtered based on a confidence threshold of 0.8. If no valid signal met the threshold, the template collection procedure was extended until at least one valid template was collected. If multiple templates were qualified, all of them were stored.

In the authentication stage, 10 seconds of PPG signals prior to authentication were used, assuming they belonged to the same person attempting to authenticate. If no valid signal was collected, the stage was extended until a valid signal was found. Multiple qualified signals were compared using an all-to-all comparison, with the final result determined by a majority vote.

After the PPG-based authentication session, participants wore the Apple Watch and registered a password. They then attempted authentication three times, while the experimenter stood approximately 50 centimeters away, trying to observe and memorize the password. The experimenter then attempted an attack by entering the memorized password. If either the participant or the experimenter entered the wrong password, the attempt was considered a failure. Participants were then asked to wear the Xiaomi Smart Band and perform the same task. These steps were repeated three times, with participants registering a different password for both the Apple Watch and Xiaomi Smart Band each time.

Next, an experiment was included to evaluate the cognitive load of participants. During a 5-minute video-based learning session session, participants were prompted to unlock the watch or band five times while watching. After the video, participants completed a questionnaire to rate their experience with the user study and express their preferences for the different authentication methods.

\subsection{Results}
\subsubsection{\textbf{Authentication Performance}}
\begin{figure}[htbp]
    \centering
    \includegraphics[width=\linewidth]{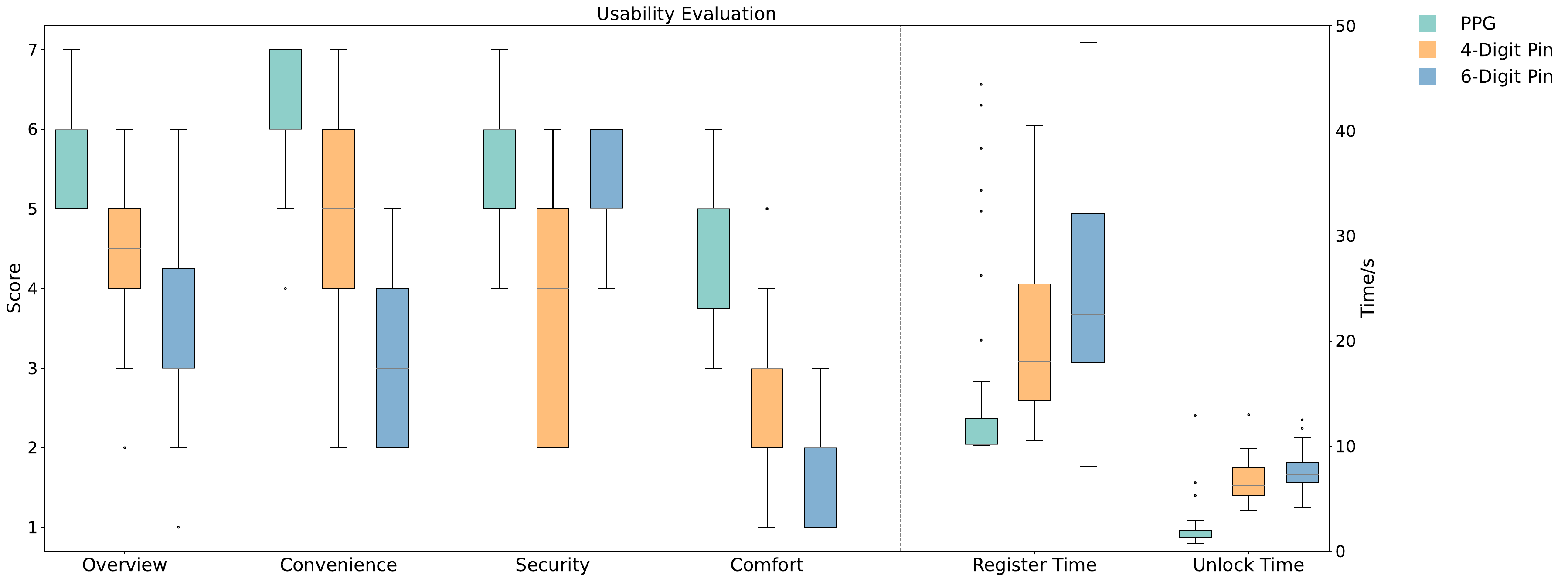}
    \caption{\textbf{Usability Evaluation.} The user preference regarding convenience, security, comfort and time consumption.}
    \label{fig: Usability Evaluation}
\end{figure}

In the experiment, we recorded several metrics for each participant, including the number of successful and failed authentication attempts, the number of successful and failed attacks by unauthorized users, and the time taken for each phase. For PPG-based smartwatches, the recorded authentication time excludes the pre-collected 10 seconds, as it is not part of the user’s intentional unlocking action. Figure \ref{fig: Usability Evaluation} presents the experimental results, showing the distribution of registration and unlock time.

The average registration time for PPG-based smartwatches was 14.7 seconds (SD=9.52s), significantly faster than that of the Apple Watch (24.8 seconds, SD=23.0s) and the Xiaomi Smart Band (28.8 seconds, SD=18.7s). The longer registration times for the Apple Watch and Xiaomi Smart Band stem from several factors.

First, their registration processes were unintuitive, with many participants failing to notice that they needed to unlock the device with an existing password before setting a new one. This led to confusion and, in some cases, device lockouts. Second, the small screen of the devices worsened usability challenges. This was particularly evident for the Xiaomi Smart Band, which featured a compact screen with 11 densely packed keys, resulting in a high likelihood of accidental touches.

The average authentication time for the PPG-based smartwatch was 1.98 seconds (SD=3.15s), significantly faster than that of the Apple Watch (6.68 seconds, SD=2.44s) and the Xiaomi Smart Band (7.55 seconds, SD=2.27s). It is worth noting that as participants became more familiar with the experimental procedures, they registered and entered PINs faster. Consequently, the first time required for PIN-based registration and authentication in real-world scenarios is likely longer than the average times reported in this study.

Regarding accuracy, the PPG-based smartwatch achieved an average authentication success rate of 98.6\% and an average attack success rate of 1.4\%, outperforming both the Apple Watch (authentication success rate = 95.8\%, attack success rate = 71.5\%) and the Xiaomi Smart Band (authentication success rate = 88.2\%, attack success rate = 16.0\%).

The Apple Watch has a high authentication success rate but suffers from a high attack success rate, likely due to its 4-digit PIN and larger screen, which reduce input errors but increase the risk of PIN observation. In contrast, the Xiaomi Smart Band, with a 6-digit PIN and smaller screen, is less prone to PIN observation from a distance but has a higher chance of input errors, explaining its lower authentication success rate and much lower attack success rate. Our PPG watch strikes a good balance between high authentication success and resistance to attacks.

\subsubsection{\textbf{User Preference}}

At the end of the experiment, we collected user experience data from all participants through a survey. The survey used a 7-point Likert scale to assess the overall experience with the three smartwatches and user experiences during key stages of the experiment. The user experience data from the 16 participants were analyzed using the Friedman test (non-parametric test), as the data distribution's normality was uncertain. Pairwise comparisons were conducted using the Wilcoxon signed-rank test, with Bonferroni correction applied to control for Type I errors.

\begin{enumerate}
    \item {\textbf{Overall experience:}
    The survey results showed the following overall experience ratings: PPG watch (M = 5.75, SD = 0.577), Apple Watch (M = 4.38, SD = 1.09), and Xiaomi Smart Band (M = 3.56, SD = 1.26). The Friedman test revealed a significant difference in the overall user experience across the three smartwatches ($X^2(2, N=16)=18.73, p<0.001$). Further Wilcoxon pairwise comparisons indicated significant differences between the PPG watch and both the Apple Watch (p = 0.027) and the Xiaomi Smart Band (p < 0.001). These results suggest that the PPG watch provided a significantly better overall user experience compared to the PIN-based devices.
    }

\item {\textbf{Convenience of authentication:}
    The survey results indicated the following convenience ratings:: PPG watch (M = 6.13, SD = 0.885), Apple Watch (M = 4.94, SD = 1.18), and Xiaomi Smart Band (M = 3.25, SD = 1.13). The Friedman test showed a significant difference in authentication convenience between the devices ($X^2(2, N=16)=23.41, p<0.001$). Wilcoxon pairwise comparisons showed significant differences between the PPG watch and both the Apple Watch (p = 0.042) and the Xiaomi Smart Band (p < 0.001). These results suggest that PPG-based authentication is significantly more convenient than PIN-based methods.
}

\item {\textbf{Security of authentication}:
    The survey results on perceived security were as follows: PPG watch (M = 5.62, SD = 0.957), Xiaomi Smart Band (M = 5.13, SD = 0.719), and Apple Watch (M = 3.69, SD = 1.44). The Friedman test revealed a significant difference in perceived security across the devices ($X^2(2, N=16)=14.37, p<0.001$). Wilcoxon pairwise comparisons showed significant differences between the PPG watch and the Apple Watch (p = 0.002). However, there was no significant difference between the PPG watch and the Xiaomi Smart Band (p = 1.00). These results indicate that users perceive 4-digit PIN (Apple Watch) as less secure than PPG and 6-digit PIN (Xiaomi Smart Band), while PPG and 6-digit PIN are considered equally secure.
}

\item {
    \textbf{Cognitive comfort}: The survey results showed that the cognitive comfort perceived by users during the video-based learning task was as follows: PPG watch (M = 4.38, SD = 0.957), Apple Watch (M = 2.88, SD = 1.088), and Xiaomi Smart Band (M = 1.81, SD = 0.75). The Friedman test showed a significant difference in cognitive load across the three devices ($X^2(2, N=16)=23.72, p<0.001$). Further Wilcoxon pairwise comparisons revealed a significant difference between the PPG watch and the Xiaomi Smart Band (p = 0.000), but no significant difference between the PPG watch and the Apple Watch (p = 0.065). These results suggest that users felt more comfortable using the PPG watch and Apple Watch.
    }
\end{enumerate}


%% file: Sections/7-Discussion.tex
\section{Discussion and Future Work}
This paper explores efficient and reliable PPG authentication on smartwatches for daily scenarios. In this section, we discuss the major results, findings, limitations and future work.

\subsection{Reliable PPG Authentication on Smartwatches}
In this paper, we demonstrated that PPG authentication faces reliability challenges due to motion artifacts and physiological variability over time. The proposed foundational RAPID block is designed based on optical physiological principles, enabling robust handling of external noise. In the static scenario with ANT\_Maxim, the RAPID block achieves a high AUC of 0.98, outperforming all baseline models. MTL-RAPID demonstrated strong performance against motion artifacts, as shown in Table~\ref{tab: motion data}. It is the state-of-the-art model compared to sequentially connected baseline models, as depicted in Figure~\ref{fig: MTL Maxim}-~\ref{fig: MTL Goodix}, achieving an AUC of 99.2\% and an EER of 3.5\% on the ANT motion dataset.

However, addressing physiological variability over time remains a key challenge for PPG-based authentication, with significant potential for improvement. To our knowledge, we are the first to explore physiological variability in the context of long-term PPG authentication. This paper presents our exploration of the MTL-RAPID model over time, which achieved a 7\% improvement in AUC on cross-day tasks compared to the best baseline. The RAPID series' consistent superiority highlights the effectiveness of its design, particularly the multi-task learning strategy.

\subsection{Efficient PPG Authentication on Smartwatches}

Authentication is a frequently used function on smartwatches, making edge-efficient methods essential for practical deployment. MTL-RAPID demonstrates significant efficiency advantages over previous methods in terms of model size, inference time, and memory usage. With only 80k parameters, MTL-RAPID is smaller than any of the models in related work. Its architecture leverages an efficient multi-task learning (MTL) design, which further reduces the parameter count. 

Additionally, MTL-RAPID offers highly efficient inference times, averaging just \textbf{1.58 ms} per operation—less than half the time of the next best-performing model, InceptionTime \cite{ismail2020inceptiontime}, which averages 3.08 ms. This substantial reduction in processing time makes MTL-RAPID a suitable solution for real-time applications, where quick response times are critical. In terms of memory usage, MTL-RAPID excels with a peak memory usage of 3.40 MB, which is only $1/10$ of InceptionTime’s\cite{ismail2020inceptiontime} 34.35 MB. This substantial reduction in memory usage makes MTL-RAPID particularly well-suited for resource-constrained devices like smartwatches or other wearables.

\subsection{Opensource Efforts for PPG authentication}
We opensource the largest wrist PPG authentication dataset ANT alongside our MTL-RAPID model to advance research in PPG-based authentication and foster collaboration. The former largest wrist PPG dataset DaLiA ~\cite{misc_ppg-dalia_495} only has a single channel signal from only 15 participants. Our ANT dataset has 30 subjects covering 3-channel PPG signals and 10 daily activities. By sharing this data, we aim to provide researchers with valuable resources to further explore the challenges in real-world PPG authentication, such as motion artifacts and physiological variability over time.

\subsection{Limits and Future Work}
We acknowledge that the size and diversity of our participant pool may not fully represent real-world demographics, and the 4-day interval may not capture the full complexity of physiological signal variability. In future work, we plan to expand the dataset to include a broader demographic and conduct longitudinal studies over longer periods. These efforts are essential for understanding PPG signal dynamics and improving the reliability of authentication systems.

As shown in Figure ~\ref{fig: MTL 1 day}-~\ref{fig: MTL 4 day}, all methods exhibit a significant decline in performance over time. This decline highlights the influence of physiological changes, such as heart rate, skin temperature, and vascular characteristics, on PPG signal consistency. Future research is needed to develop reliable solutions for addressing physiological variability. Potential directions include adaptive learning techniques to continuously fine-tune the model based on evolving physiological patterns, as well as multi-modal authentication systems combining PPG with other biometric signals, such as accelerometer data.

We recommend adopting an updated template strategy~\cite{rattani2009template} to better accommodate daily variations in PPG signals. We argue that temporal changes in PPG signals could enhance theft prevention, albeit posing challenges for algorithmic performance. Additionally, developing advanced algorithms to compensate for physiological variability could maintain high authentication accuracy over extended periods, improving the viability of PPG-based authentication in real-world applications.